\documentclass[10pt,letterpaper]{article}

\usepackage{dirtytalk}
\usepackage{amsfonts}

\usepackage{soul}
\soulregister{\cite}{1}
\soulregister{\ref}{1}

\usepackage[title]{appendix}

\makeatletter
\newcommand\subsubsubsection{\@startsection{paragraph}{4}{\z@}%
  {-3.25ex \@plus -1ex \@minus -.2ex}%
  {1.5ex \@plus .2ex}%
  {\normalfont\normalsize\bfseries}}
\makeatother

\usepackage{caption}
\usepackage[utf8]{inputenc}

\usepackage{hhline}
\usepackage{makecell}
\usepackage{subcaption}
\usepackage{fullpage}
\usepackage{times}
\usepackage{array}
\usepackage{ragged2e}
\usepackage{multirow}
\usepackage{fancyhdr, amssymb}
\usepackage[ruled,vlined]{algorithm2e}
\usepackage[margin = 1 in]{geometry}
\usepackage{graphicx}
\usepackage{outlines}
\usepackage{longtable}
\usepackage{amsmath}

\usepackage{url}
\usepackage[english]{babel} 
\usepackage{blindtext}
\usepackage{xltabular}
\usepackage{arydshln} 
\usepackage{wrapfig}
\usepackage{enumitem}
\usepackage{floatrow}
\usepackage{xfrac}
\usepackage[font={footnotesize}, labelfont={bf}, labelsep=space, justification=justified, skip=0pt]{caption}
\usepackage[hang,flushmargin]{footmisc} 
\usepackage{fancyhdr, amssymb, mathtools}
\usepackage{csquotes}
\usepackage{amssymb}
\usepackage{bbm}
\usepackage{mathtools}
\usepackage{perpage}

\usepackage[dvipsnames,table]{xcolor}
\usepackage[
backend=biber,
style=numeric-comp,
sorting=none,
maxnames = 10
]{biblatex}
\newcommand{\gp}{\texttt{GP+}}
\newcommand{\mfbo}{\texttt{MFBO}}
\newcommand{\cmfbo}{\texttt{CMFBO}}
\newcommand{\csfbo}{\texttt{CSFBO}}

\newcommand{\betab}{\boldsymbol{\beta}}

\newcommand{\thetab}{\boldsymbol{\theta}}

\newcommand{\omegab}{\boldsymbol{\omega}}

\newcommand\braces[1]{\mathopen{}\left\{#1\right\}\mathclose{}}
\newcommand{\xsspace}{\mathbb{X}}
\newcommand{\rsspace}{\mathbb{R}}

\DeclareMathOperator*{\E}{\mathbb{E}}

\newcommand{\yb}{\boldsymbol{y}}

\newcommand{\ub}{\boldsymbol{u}}
\newcommand{\Ub}{\boldsymbol{U}}
\newcommand{\tb}{\boldsymbol{t}}

\newcommand{\xb}{\boldsymbol{x}}
\newcommand{\Xb}{\boldsymbol{X}}
\newcommand{\Ib}{\boldsymbol{I}}
\newcommand{\zb}{\boldsymbol{z}}
\newcommand{\hb}{\boldsymbol{h}}
\newcommand{\mb}{\boldsymbol{m}}

\newcommand{\Cb}{\boldsymbol{C}}

\newcommand{\pib}{\boldsymbol{\pi}}

\newcommand{\wing}{\texttt{Wing}}
\newcommand{\sepwing}{$\texttt{Wing}_\texttt{Sep}$}

\newcommand{\braninh}{\texttt{Branin-Hoo}}
\newcommand{\hartman}{\texttt{Hartman}}
\newcommand{\twenty}{\texttt{PolyMix}}

\usepackage[colorlinks=true, linkcolor=blue, citecolor=blue, filecolor=blue, urlcolor=blue]{hyperref}
\usepackage{cleveref}

\AtBeginEnvironment{appendices}{\crefalias{section}{appendix}}

\makeatletter
\AtBeginDocument{%
  \g@addto@macro\normalsize{%
    \setlength\abovedisplayskip{8pt}
    \setlength\belowdisplayskip{8pt}
    \setlength\abovedisplayshortskip{8pt}
    \setlength\belowdisplayshortskip{8pt}
    \let\orig@setfontsize\@setfontsize
  }%
}
\makeatother

\DeclareMathSizes{10}{10}{7}{6}

\newcommand{\cmt}[1]{} 

{\title{\fontsize{14}{14}\selectfont

CONSTRAINED MULTI-FIDELITY BAYESIAN OPTIMIZATION WITH AUTOMATIC STOP CONDITION

}}

\date{\vspace{-5ex}}
\usepackage{authblk}
\author[1]{Zahra Zanjani Foumani}
\author[1,2]{Ramin Bostanabad\thanks{\noindent Corresponding Author: raminb@uci.edu}}

\affil[1]{Department of Mechanical and Aerospace Engineering, University of California, Irvine, CA 92697, USA.}

\affil[2]{Department of Civil and Environmental Engineering, University of California, Irvine, CA 92697, USA.}

\begin{document}
    \include{pythonlisting}
    \pagenumbering{arabic}
    \sloppy
    \maketitle
    \begin{abstract} 
Bayesian optimization (BO) is increasingly employed in critical applications to find the optimal design with minimal cost. While BO is known for its sample efficiency, relying solely on costly high-fidelity data can still result in high costs. This is especially the case in constrained search spaces where BO must not only optimize but also ensure feasibility. A related issue in the BO literature is the lack of a systematic stopping criterion. 
To solve these challenges, we develop a constrained cost-aware multi-fidelity BO (\cmfbo) framework whose goal is to minimize overall sampling costs by utilizing inexpensive low-fidelity sources while ensuring feasibility. In our case, the constraints can change across the data sources and may be even black-box functions. We also introduce a systematic stopping criterion that addresses the long-lasting issue associated with BO's convergence assessment. Our framework is publicly available on GitHub through the \gp~Python package and herein we validate it's efficacy on multiple benchmark problems.
\end{abstract}
    \section{Introduction} \label{sec: intro}
Bayesian Optimization (BO) is a sequential and sample-efficient global optimization technique widely used for optimizing expensive-to-evaluate, black-box functions \cite{brochu2010tutorial}. BO has two main components: an acquisition function (AF) and an emulator which is often a Gaussian Process (GP). Starting with an initial dataset, BO iteratively refines the emulator by adding new samples using the AF until convergence criteria are satisfied \cite{couckuyt2022bayesian}.

Despite its efficiency, BO can be costly when relying solely on accurate and expensive high-fidelity (HF) data. To mitigate this, multi-fidelity (MF) techniques have been developed \cite{takeno2020multi} to leverage data sources of varying accuracy and cost levels to reduce overall sampling expenses by exploiting the correlations between HF and low-fidelity (LF) datasets. Compared to single-fidelity BO (SFBO), the choice of emulator in MFBO is more critical due to the heterogeneous nature of the data. In MF settings, GPs are still favored over alternatives like probabilistic neural networks which are costly to train and prone to overfit in data-scarce regimes \cite{li2022line,moslemi2023scaling,ghorbanian2024empowering}.

Numerous MFBO strategies have been recently proposed but they mostly struggle to design appropriate emulators and/or AFs to handle the MF nature of the problem.
For instance, many MF methods require prior knowledge about the hierarchy of LF data sources \cite{picheny2013quantile}. Some modeling approaches train separate emulators for each data source and hence fail to fully leverage cross-source correlations \cite{sun2022correlated,lam2015multifidelity}. Popular bi-fidelity approaches based on Kennedy and O’Hagan's work and its extensions \cite{kennedy2001bayesian} are limited to two fidelities and often assume simple bias forms (e.g., additive functions) for LF sources \cite{olleak2020calibration}. Similarly, Co-Kriging and its extensions \cite{zhou2020generalized, chen2023topsis,shi2020multi} often fail to accurately model cross-fidelity relationships. 
The limitations on MF emulation are typically exacerbated by the AF as it may encourage over-sampling of very cheap LF sources. Such a behavior leads to inaccurate results and affects even popular packages like \texttt{botorch} \cite{gardner2018gpytorch}.

Existing BO methods are primarily designed for unconstrained problems. While there are a few constrained strategies available, they are mostly limited to SF cases \cite{csf1,csf2,schonlau1998global,csf4}. A few MFBO methods address constrained problems but they either suffer from the same emulation limitations discussed above \cite{tran2019sbf,lin2023multi} or are restricted to bi-fidelity settings \cite{zhang2018variable,shu2021multi,tran2020smf}.

Most constrained MFBO methods rely on the Probability of Feasibility (PoF) to account for feasibility in their AFs \cite{pof1,shu2021multi,ur2017expected,tran2022aphbo}. These methods require a feasibility threshold which cannot be chosen systematically and is sensitive to noisy data.
Additionally, PoF requires an initially feasible solution to proceed \cite{zhou2024multi,he2022variable} which is an assumption that is regularly violated in high-dimensional spaces. Additionally, PoF does not consider the severity of constraint violations and hence fails to direct the search towards feasible designs. 
Although some constrained MFBO methods don't rely on PoF, they also have some limitations. Many assume identical constraints across fidelity levels but LF sources typically fail to provide an accurate representation of the HF source's constraints 
\cite{di2023nm,sorourifar2023computationally,he2022batched}. Others adopt multi-step look-ahead frameworks to improve feasibility, but this comes at the cost of increased complexity and computational overhead \cite{ghoreishi2019multi,khatamsaz2023bayesian}.

The stop condition is a critical component of all BO methods including MF, constrained, and other versions.
In iterative processes like BO, defining a proper stopping criterion is essential to balance cost and solution quality. Too many iterations waste resources while stopping prematurely risks suboptimal solutions. Despite its importance, this topic has received limited attention. Most approaches use simple and problem-dependent metrics such as maximum number of iterations or improvement threshold \cite{berkenkamp2023bayesian,wang2024constrained,folch2024transition}.

Our recent work, \mfbo~\cite{zanjani2024safeguarding}, addresses most of the above limitations in two ways. First, it designs GPs \cite{eweis2022data} that can simultaneously fuse any number of data sources without imposing any constraint or assumption on the hierarchy of the data sources (e.g., linear correlation among sources, additive biases, noise level, etc.). Second, it quantifies the information value of LF and HF samples differently to consider the MF nature of the data while exploring the search space \cite{foumani2023multi}. Specifically, the AF used in \mfbo~is cost-aware in that it considers the sampling cost in quantifying the value of HF and LF data points. 
While \mfbo~performs well, it shares two key limitations with other MFBO methods: $(1)$ the inability to handle constraints, $(2)$ the lack of systematic stopping criteria. We introduce \cmfbo~ in this paper to address these issues.

The rest of the paper is organized as follows. We introduce our approach in Section \ref{sec: Method} and then evaluate its performance on multiple benchmark problems in Section \ref{sec: results}. We conclude the paper in Section \ref{sec: conclusion} by summarizing our contributions and providing future research directions.

    \section{Methods} \label{sec: Method}
We begin by providing background on MF modeling with GPs in Section \ref{sec: back_GP} and then introduce our novel AF for handling constraints in Section \ref{sec: AFs}. Finally, in Section \ref{sec: stop} we develop a stop condition that automatically adapts to the problem.
Our approach is implemented via the open-source Python package \gp~\cite{gp+}.

\subsection{Emulation with Gaussian Processes (GPs)}\label{sec: back_GP}
We first review GPs and their extension to handle mixed-input domains which seamlessly enables MF emulation. Then, we introduce a few mechanisms to improve the uncertainty quantification (UQ) accuracy of our emulator which directly benefits MFBO.

\subsubsection{Mixed-Input Emulation via GPs} \label{sec: vanilla_GPs} 
GPs are probabilistic models that assume the training data follows a multivariate normal distribution whose mean vector and covariance matrix depend on the inputs.
Assume the training dataset $\braces{\xb^{(i)}, y^{(i)}}_{i=1}^n$ is given where $\xb = [x_1, ..., x_{dx}] \in \xsspace \subset \rsspace^{dx}$ and $y^{(i)} = y(\xb^{(i)}) \in \rsspace$ denote the inputs and response, respectively\footnote{Superscript numbers enclosed in parenthesis indicate sample numbers. For instance, $x^{(i)}$ denotes the $i^{th}$ sample in a training dataset while $x_i$ indicates the $i^{th}$ component of the vector $\xb = [x_1, \cdots, x_{dx}]$.}. Given $\yb = [y^{(1)}, \cdots, y^{(n)}]^T$ and $\Xb$ whose $i^{th}$ row is $\xb^{(i)}$, our goal is to predict $y(\xb^*)$ at the arbitrary point $\xb^* \in \xsspace$.
Following the above description, we assume $\yb$ is a realization of a GP with the following parametric mean and covariance functions:
\begin{subequations} 
    \begin{equation} 
        \mathbb{E}[y(\xb)] = m(\xb; \betab),
        \label{eq: gp-mean}
    \end{equation}
    \begin{equation} 
        \text{cov}\left(y(\xb), y(\xb')\right) = c(\xb, \xb'; \sigma^2, \thetab) = \sigma^2 r(\xb, \xb'; \thetab)
        \label{eq: gp-cov}
    \end{equation}
    \label{eq: gp-mean and cov}
\end{subequations} 
\noindent where $\betab$, $\sigma^2$, and $\thetab$ are the parameters of the mean and covariance functions. The mean function $m(\xb; \betab)$ can take various forms such as polynomials or neural networks (NNs). In practice, a constant mean function, $m(\xb; \betab) = \beta$, is often used which causes the GP's performance to heavily rely on its kernel.

The covariance function or kernel in Eq.\ref{eq: gp-cov} is formulated by scaling the correlation function $r(\xb, \xb'; \thetab)$ parameterized by $\thetab$ via the process variance $\sigma^2$. Common choices for $r(\cdot, \cdot)$ include Gaussian, power exponential, and Matérn kernels. In this work, we employ the Gaussian kernel:
\begin{equation} 
    r\left(\xb, \xb^{\prime}; \omegab \right) = 
            \exp \left\{-\sum_{i=1}^{dx} 10^{\omega_i}(x_i-x_i^{\prime})^2 \right \}  
    \label{eq: rbf-kernel}
\end{equation}
\noindent where $\omega=\thetab$ are the kernel parameters.

The kernel in Eq.\ref{eq: rbf-kernel} fails to handle categorical features which are not naturally endowed with a distance metric. To address this issue, we use the idea proposed in \cite{oune2021latent} and introduce quantitative prior representations $\pib_t = f_{\pi}(\tb)$ for categorical variables $\tb = [t_1, \ldots, t_{dt}]$, where $f_{\pi}(\cdot)$ is a user-specified deterministic function such as grouped one-hot encoding which is used in this paper. These priors are often high-dimensional ($d_\pi > d_t$) and are passed through a parametric embedding function $f_h(\pib_t; \thetab_h)$ to yield a low-dimensional latent representation $\hb = f_h(f_\pi(\tb); \thetab_h)$, where $d_\pi \gg d_h$. Incorporating $\hb$ into Eq.\ref{eq: rbf-kernel} we obtain:
\begin{equation} 
    r(\ub, \ub^{\prime}; \omegab, \thetab_h)= 
        \exp \left\{-\sum_{i=1}^{dx} 10^{\omega_i}(x_i-x_i^{\prime})^2\ -
        \sum_{i=1}^{dh}(h_i-h_i^{\prime})^2 \right\}
    \label{eq: matern-kernel-GP_Plus}
\end{equation}
\noindent where $\ub = \begin{bmatrix} \xb , \tb \end{bmatrix}$ and $\thetab = \{\omegab, \thetab_h\}$. 

To estimate the parameters of a GP, we leverage maximum likelihood estimation (MLE) which is equivalent to minimizing the negative of the marginal likelihood:
\begin{equation} 
    \begin{split}
        [\widehat{\betab}, \widehat{\sigma}^2, \widehat{\thetab}] 
        &= \underset{\betab, \sigma^2, \thetab}{\operatorname{argmin}} \hspace{2mm} L_{MLE} \\
        &= \underset{\betab, \sigma^2, \thetab}{\operatorname{argmin}} \hspace{2mm} 
        \frac{1}{2} \log (|\Cb|) 
        + \frac{1}{2} (\yb - \mb)^T \Cb^{-1} (\yb - \mb)
    \end{split}
    \label{eq: map-gp}
\end{equation}
\noindent where $\Cb$ is the covariance matrix with entries $C_{ij} = c(\ub^{(i)}, \ub^{(j)}; \sigma^2, \thetab)$ and $\mb$ is an $n \times 1$ vector whose $i^{th}$ element is $m_i=m(\ub^{(i)}; \betab)$. Once the parameters are estimated, predictions at a new point $\ub^*$ are obtained using:
\begin{subequations} 
    \begin{equation} 
        \E[y(\ub^*)] = \mu(\ub^*) = 
        m(\ub^*; \widehat{\betab}) + c(\ub^*, \Ub; \widehat{\thetab}, \widehat{\sigma}^2) \boldsymbol{C}^{-1}(\yb-\mb)
        \label{eq: gp-mean-scalar}
    \end{equation}
    \begin{equation}
    \begin{split}
        \text{cov}(y(\ub^*), y(\ub^*)) &= \tau^2(\ub^*) = c(\ub^*, \ub^*; \widehat{\thetab}, \widehat{\sigma}^2) \\
        &\quad - c(\ub^*, \Ub; \widehat{\thetab}, \widehat{\sigma}^2) \Cb^{-1} 
        c(\Ub, \ub^*; \widehat{\thetab}, \widehat{\sigma}^2)
    \end{split}
    \label{eq: gp-var-scalar}
\end{equation}
    \label{eq: gp-mean-var-scalar}
\end{subequations}    
\noindent where  $c(\ub^*, \Ub; \widehat{\thetab}, \widehat{\sigma}^2)$ represents the covariance between $\ub^*$ and the training points.

\subsubsection{Multi-fidelity Modeling with GPs} \label{sec: MF_emulation}
\begin{figure}[!b] 
    \centering
    \includegraphics[page=1, width = 0.47\textwidth]{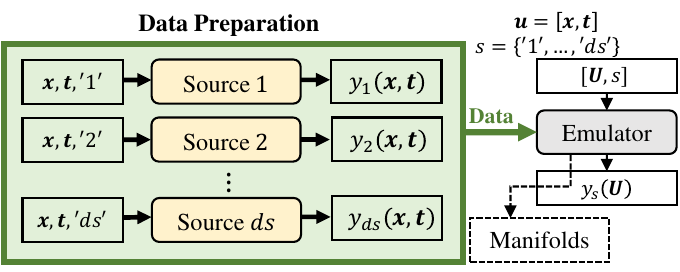}
    \vspace{-0.3cm}
    \caption{{\textbf{MF Modeling with GPs:}} The training data is built by first augmenting the inputs with the \textit{categorical} feature $s$ and then concatenating all the inputs and outputs. For MF emulation, we recommend using two manifolds to simplify the visualization of cross-source relationships: one manifold for $s$ and the other for the rest of the categorical variables, i.e., $\boldsymbol{t}$.}
    \label{fig: LMGP-flowchart}
\end{figure}

As schematically illustrated in Fig.\ref{fig: LMGP-flowchart}, the first step to MF emulation using \cite{oune2021latent} is to augment the inputs with the additional \textit{categorical} variable $s$ that indicates the source of a sample, i.e., $s=\{'1','2',\ \ldots,'ds'\}$ where the $j^{th}$ element corresponds to source $j$ for $j=1,\ldots,ds$. Subsequently, the training data from all sources are concatenated to build an MF emulator. Upon training, to predict the objective function value at $\begin{bmatrix} \xb^* ,\tb^* \end{bmatrix}$ from source $j$, the query point is concatenated with the categorical variable $s$ that corresponds to source $j$ and fed into the GP. 

We refer the readers to \cite{eweis2022data,gp+,oune2021latent} for more detail but note here that in case the input variables already contain some categorical features ($\tb$~in Fig.\ref{fig: LMGP-flowchart}), by default \gp~learns two different manifolds where one encodes the fidelity variable $s$ while the other manifold encodes the rest of the categorical variables:
\begin{equation}
    \begin{split}
        r([\ub,s], [\ub^{\prime},s^{\prime}]; \omegab, \thetab_h, \thetab_z) = 
        \exp \Bigg\{ 
            - \sum_{i=1}^{dx} 10^{\omega_i} (x_i - x_i^{\prime})^2 \\
            - \sum_{i=1}^{dh} (h_i - h_i^{\prime})^2 
            - \sum_{i=1}^{dz} (z_i - z_i^{\prime})^2
        \Bigg\}
    \end{split}
    \label{eq: rbf-kernel-GP_Plus-mf}
\end{equation}
\noindent where $\ub= \left[ \xb, \tb \right]$ and $\boldsymbol{z} = f_z(\boldsymbol{\pi}_s; \thetab_z)$ is the latent representation of data source $s$ and is obtained similar to $\hb$.

\subsubsection{Source-Dependent Noise Modeling} \label{sec: multi_noise}
Noise is an inherent feature in most applications and improperly modeling it can significantly degrade the performance of any emulator. GPs model noise using the nugget (or jitter) parameter, $\delta$, which modifies the covariance matrix from $\boldsymbol{C}$ to $\Cb_{\delta} = \Cb + \delta \Ib_{nn}$ \cite{bostanabad2018leveraging}. While this approach performs well for SF problems, it proves inadequate for MF emulation due to the fundamentally different characteristics of the data sources and their associated noise.
To address this limitation, we adopt the approach outlined in \cite{zanjani2024safeguarding}, employing a nugget vector $\boldsymbol{\delta} = [\delta_1, \delta_2, \dots, \delta_{d_s}]$ to refine the covariance matrix as follows:
\begin{equation} 
    \boldsymbol{C}_\delta=\boldsymbol{C}+\boldsymbol{N}_\delta
    \label{eq: sep_noise_corr}
\end{equation}
\noindent where $\boldsymbol{N}_\delta$ denotes an $n \times n$ diagonal matrix whose $(i,i)^{th}$ element is the nugget element corresponding to the data source of the $i^{th}$  sample. 

\subsubsection{Emulation for Exploration} \label{sec: Interval_score}
Proper UQ plays a critical role in MFBO, as it directly affects the exploration behavior of the AF and the selection of subsequent samples. While source-dependent noise modeling improves the emulator’s ability to capture uncertainty by introducing additional hyperparameters, it may also lead to overfitting \cite{mohammed2017over}. Furthermore, large local biases in LF sources can inflate uncertainties, disrupt AF exploration, and potentially cause convergence to suboptimal solutions.

To address these challenges, we follow \cite{zanjani2024safeguarding} and use strictly proper scoring rules like the negatively oriented interval score ($IS$) to emphasize UQ during training. $IS$ is robust to outliers and rewards (penalizes) narrow prediction intervals that contain (miss) the training samples \cite{bracher2021evaluating}. For a GP, $IS$ can be analytically calculated for $(1-v) \times 100 \%$ prediction interval as:
\begin{equation}
        IS_{\nu} = \frac{1}{n} \sum_{i=1}^n (\mathcal{U}_i - \mathcal{L}_i)
        + \frac{2}{\nu} (\mathcal{L}_i - y_j(\boldsymbol{u}^{(i)})) 
        \mathbbm{1}\{y_j(\boldsymbol{u}^{(i)}) < \mathcal{L}_i\} 
        + \frac{2}{\nu} (y_j(\boldsymbol{u}^{(i)}) - \mathcal{U}^{(i)}) 
        \mathbbm{1}\{y_j(\boldsymbol{u}^{(i)}) > \mathcal{U}_i\}
    \label{eq: IS}
\end{equation}
\noindent where $\mathbbm{1}\{\cdot\}$ is an indicator function which is $1$ if its condition holds and zero otherwise. We use $\nu=0.05$  ($95\%$ prediction interval), so $\mathcal{U}_i= \mu_j{(\boldsymbol{u}^{(i)})}+1.96 \tau_j{(\boldsymbol{u}^{(i)} )}$ and $\mathcal{L}_i= \mu_j{(\boldsymbol{u}^{(i)} )}-1.96 \tau_j{(\boldsymbol{u}^{(i)})}$. 

To incorporate $IS$ into training, we scale it to ensure it has a comparable magnitude to $L_{MLE}$ and add it to Eq.\ref{eq: map-gp}:
\begin{equation} 
     {[\hat{\beta}, \widehat{\sigma}^2,\widehat{\thetab} ]=}  
     \underset{\beta, \sigma^2, \thetab}{\operatorname{argmin}} \hspace{3mm} L_{MLE}  +\varepsilon|L_{MLE}| \times IS_{0.05}
    \label{eq: IS_based_objective}
\end{equation}
\noindent where $|\cdot|$ denotes the absolute function and $\varepsilon$ is a user-defined scaling parameter (we use $\varepsilon=0.08$ for all of our examples in this paper). 

\subsubsection{Emulation with Mixed Basis Functions} \label{sec: gp_mixed_basis_functions}
The parametric mean and covariance functions in Eq.\ref{eq: gp-mean and cov} can be formulated in many ways. While most advancements have focused on designing the kernel, the mean function in Eq.\ref{eq: gp-mean} plays an important role in many applications, especially in MF modeling.

\begin{figure*}[t] 
    \centering
    \includegraphics[width=\linewidth]{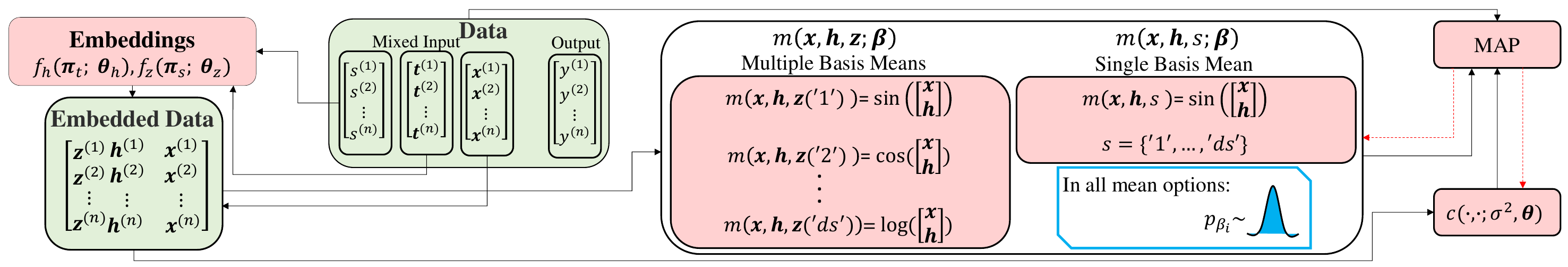}
    \caption{\textbf{Mixed vs Single Basis Functions:} Two generic options are defined in \gp~for building the mean function: $(1)$ mixed basis where multiple bases can be defined for each data source,
    $(2)$ single basis where a global shared function is learned for all the data sources.}
    \label{fig: Mixed_Basis_Functions}
\end{figure*}

In our approach, we feed the learnt representations of $\tb$ and $s$ into the mean function instead of the original categorical variables and reparameterize $m(\xb, \tb, s; \betab)$ to $m(\xb, \hb, \zb; \betab)$ \cite{gp+}, see Fig.\ref{fig: Mixed_Basis_Functions}. Following this reformulation, we can use mixed basis functions where a unique mean function is learnt for specific combinations of the categorical variables (e.g., in MF modeling, we can learn a unique mean function for each of the $s$ data sources) which can be advantageous to learning a global mean function for \textit{all} combinations of $\tb$ and $s$.

\subsection{Proposed Acquisition Function} \label{sec: AFs}
The choice of AF is crucial in BO as it balances exploration and exploitation during sampling. In MFBO, this choice becomes even more critical, as the AF must also account for biases in LF data and source-dependent sampling costs. Additionally, when constraints are present, the AF must ensure feasibility while optimizing information gain and cost. 

To handle the biases of the LF sources and balance information gain and cost, separate AFs are defined in \mfbo~for LF and HF sources where the cheap LF sources are primarily used for exploration while the expensive HF samples are maximally exploited. 
Following this idea, the AF of the $j^{th}$ LF source ($j \neq l$, $l$ denotes the HF source) is defined as the exploration part of the expected improvement (EI) in \mfbo:
\begin{equation} 
    \gamma_{LF}(\boldsymbol{u} ; j)=\tau_j(\boldsymbol{u}) \phi(\frac{y_j^*-\mu_j(\boldsymbol{u})}{\tau_j(\boldsymbol{u})})
    \label{eq: AF_LF}
\end{equation}
\noindent where $y_j^*$ is the best (potentially noisy) function value obtained so far from source $j$ and $\phi(\cdot)$ denotes the probability density function (PDF) of the standard normal variable. $\tau_j(\boldsymbol{u})$ and $\mu_j(\boldsymbol{u})$ are the standard deviation and mean, respectively, of point $\boldsymbol{u}$ from source $j$ which we estimate via Eq.\ref{eq: gp-mean-var-scalar}.

\mfbo~utilizes improvement as the AF for the HF data source since it is computationally efficient and emphasizes exploitation:
\begin{equation} 
    \gamma_{HF}(\boldsymbol{u} ; l)=y_l^*-\mu_l(\boldsymbol{u})
    \label{eq: HF_AF}
\end{equation}

We extend these AFs in \cmfbo~to incorporate constraints which are either known or unknown. While the former can be evaluated analytically, the latter are more complex and require costly simulations. Therefore, we specifically focus on unknown constraints and note that the proposed approach can easily handle known ones in a similar way. 

The optimization problem with unknown constraints is formulated as follows:
\[
\min_{\ub} f_l(\ub) \quad \text{subject to} \quad G^{k}_l(\ub) \leq 0, \; k = 1, \dots, K.
\]
\noindent where $f_l(\ub)$ is the HF black-box function we aim to optimize and $G^{k}_l(\ub)$ denotes the $k^{th}$ unknown constraint corresponding to the HF source. 

The unknown constraints are black-box functions observed alongside \( y_j(\ub) \). So, in constrained MFBO, in addition to the function value \( y_j(\ub) \), \( K \) constraint values are also observed (\( g^{k}_j(\ub), k=1,\dots,K \)) for source \( j \).
To model \( G^{k}_j(\ub) \), we assign GP priors to them. 
Since different constraints typically behave quite differently, in our approach we learn an independent GP for constraint $k$. We fit this GP in an MF context since constraint $k$ may not be the same across all data sources. 
For example, in an application with three data sources and two constraints which depend on the fidelity of the data sources, we build three GPs: one for the objective function and one for each constraint. Each model fuses data from three fidelities as explained in Section \ref{sec: back_GP}. 

By emulating the constraints via GPs we can assign a normal distribution to the value of the $k^{th}$ constraint of the $j^{th}$ data source (that is, $g^{k}_j$). We denote the mean and variance of these predictions via \( \mu_{g^{k}_j}(\ub) \) and \( \tau^2_{g^{k}_j}(\ub) \),  respectively, and use them to update Eq.\ref{eq: AF_LF} and Eq.\ref{eq: HF_AF} to consider feasibility. 
Specifically, the AFs in  Eq.\ref{eq: AF_LF} and Eq.\ref{eq: HF_AF} are used for samples that satisfy all constraints, i.e., samples with $\mu_{g^{k}_j}(\boldsymbol{u}) \leq 0, \, \forall k = 1, \dots, K$.
However, if any constraint is violated (that is, if $\mu_{g^{k}_j}(\boldsymbol{u}) > 0 \, \text{for some } k$) the AF minimizes the violations to move toward the feasible domain (in cases with multiple violations, the summation of the violated constraints is minimized). Hence, our proposed AFs for LF and HF sources are:
\begin{subequations}
    \begin{equation}
        \gamma_{C-LF}(\boldsymbol{u}; j) =
        \begin{cases}\label{eq: AF_C_LF}
            \tau_j(\boldsymbol{u}) \phi \left(
                \frac{y_j^* - \mu_j(\boldsymbol{u})}{\tau_j(\boldsymbol{u})}
            \right)
            & \text{if } \mu_{g^{k}_j}(\boldsymbol{u}) \leq 0, \forall k\\[1em]
            -\sum_{k=1}^{K} \mu_{g^{k}_j}(\boldsymbol{u}) 
            & \text{if } \mu_{g^{k}_j}(\boldsymbol{u}) > 0 \text{ for some } k
        \end{cases}
    \end{equation}
    
    \vspace{1em}
    
    \begin{equation} \label{eq: AF_HF_c}
        \gamma_{C-HF}(\boldsymbol{u}; l) =
        \begin{cases}
            y_l^* - \mu_l(\boldsymbol{u}) 
            & \text{if } \mu_{g^{k}_l}(\boldsymbol{u}) \leq 0, \forall k \\[1em]
            -\sum_{k=1}^{K} \mu_{g^{k}_l}(\boldsymbol{u}) 
            & \text{if } \mu_{g^{k}_l}(\boldsymbol{u}) > 0 \text{ for some } k
        \end{cases}
    \end{equation}
\end{subequations}

In each iteration of BO, we first use the proposed AFs to solve $ds$ auxiliary optimizations
and find the feasible candidate points with the highest acquisition value from each source. We then scale these values by the corresponding sampling costs to obtain the following composite cost-aware AF:
\begin{equation} 
    \begin{split}
    \gamma_{\cmfbo}(\boldsymbol{u}; j) = 
    \begin{cases}
        \sfrac{\gamma_{C-LF}(\boldsymbol{u}; j)}{O(j)} & j = [1, \cdots, ds] \And j \neq l \\
        \sfrac{\gamma_{C-HF}(\boldsymbol{u}; l)}{O(l)} & j = l
    \end{cases}
    \end{split}
    \label{eq: composite_AF}
\end{equation}
\noindent where $O(j)$ is the cost of acquiring one sample from source $j$. We determine the final candidate point (and the source that it should be sampled from) at iteration $q+1$ via:
\begin{equation} 
    [\boldsymbol{u}^{(q+1)}, j^{(q+1)}]=\underset{\boldsymbol{u}, j}{\operatorname{argmax}}~ \gamma_{\cmfbo}(\boldsymbol{u} ; j)
    \label{eq: auxiliary-opt}
\end{equation}
\subsection{Proposed Convergence Metric} \label{sec: stop}
Our core idea is based on the observation that as more and more data points are sampled during BO iterations, the emulator more accurately approximates the HF source, particularly around its local optima. We argue that at \textit{some} point during the BO iterations, these incremental improvements start to provide diminishing returns, i.e., $y_l^*$ negligibly improves with further sampling from either the HF or LF sources. To identify the iteration where such a behavior emerges, we propose to track the evolution of the optima of $\mu_l(\ub)$ during BO iterations and halt the optimization once the locations of these optima stabilize.

To track the optima of $\mu_l(\ub)$ subject to $g_l^k(\ub)$ for $k=1,\dots,K$, we carry out a post-auxiliary optimization (PAO) at the end of each BO iteration. Using a gradient-based optimizer, we find the optima of $\mu_l(\ub)$, denoted by \( (\ub^*_{POA}, y^*_{POA}) \). If the identified optima negligibly change over \( v \) consecutive iterations, we conclude that the emulator is stabilized and terminate the optimization process.

To account for the scale of the problems, we normalize the optimum values found by the PAO. Specifically, we compute the mean ($\mu_{y^*_{sec}}$) and standard deviation ($\tau_{y^*_{sec}}$) of the $y^*_{sec}$ values from all previous iterations and normalize them as follows:
\begin{subequations}\label{eq: normalization_PAO}
    \begin{equation}
        \mu_{y^*_{PAO}} = \frac{1}{q} \sum_{i=1}^{q} y_{PAO}^{* (i)}
    \end{equation}
    \begin{equation}
        \quad \tau_{y^*_{PAO}} = \sqrt{\frac{1}{q} \sum_{i=1}^{q} \left(y_{PAO}^{* (i)} - \mu_{y^*_{PAO}}\right)^2}
    \end{equation}
    \begin{equation}
    \tilde{y}_{PAO}^{* (i)} = \frac{y_{PAO}^{* (i)} - \mu_{y^*_{PAO}}}{\tau_{y^*_{PAO}}}
    \end{equation}
\end{subequations}
where $\tilde{y}_{PAO}^{* (i)}$ is the normalized post-auxiliary optimum at iteration $i$. Then, the stopping criterion is defined as follows:
\begin{equation} \label{eq: threshold}
    var(\{\tilde{y}_{PAO}^{*(i)}\}_{i=q-v+1}^{q}) < \epsilon
\end{equation}
with $var$ representing variance and $\epsilon$ is a user defined threshold.

We note as the PAO also aims to minimize $f_l(\ub)$, the optimum value reported at the end of BO, $y^*_{final}$, is the smaller of two values (considering a minimization problem): either $y^*_l$ which is the minimum found using the AF or the minimum feasible optimum found in the last $v$ iterations of the PAO. That is:
\begin{equation}
    y^*_{final} = \min\{y^*_l, \min\{{y}_{sec}^{*(i)}\}_{i=q-v+1}^q\}
\end{equation}

In our implementation of PAO we use the Limited-memory Broyden–Fletcher–Goldfarb–Shanno (L-BFGS) algorithm \cite{saputro2017limited} which is a gradient-based method whose performance depends on the initialization. To reduce this dependence, we design a \textit{look-back initialization} strategy where PAO at iteration $q \geq 2$ is restarted from $(1)$ feasible HF samples collected so far, and $(2)$ the $10$ optimum points identified by PAO at iteration $q-1$ (when $q=1$, we initialize PAO with $30$ randomly generated samples alongside the feasible HF samples). This strategy reduces the computational cost of a fully random initialization while focusing the search on the most promising regions of the design space.  

We present a summary of our proposed framework (\cmfbo) in Algorithm \ref{alg: MFBO-algorithm} and conclude this section by noting that we generally use larger $\epsilon$ values for very high dimensional problems where the HF emulator can noticeably change around its local optima as more HF or LF samples are collected. For instance, for the $20D$ problem in Section \ref{sec: results} we set $\epsilon$ to $0.5$ while for all other problems in that section, we use $\epsilon=0.01$.

\begin{algorithm*}[!t]
    \SetAlgoLined
    \DontPrintSemicolon
    \textbf{Given:} Initial constrained MF data $\mathcal{D}^{q}=\{\ub^{(i)}, y_j(\ub^{(i)}), g_j^{k}(\ub^{(i)})\}_{i=1}^q$ for $j = 1, \dots, ds$ and $\forall k; k=1,\dots,K$, black-box objective functions $f_j(\ub)$ and their corresponding sampling costs $O(j)$, and black-box constraint functions $G^{k}_j(\ub)$.\\
    \textbf{Goal:} Optimizing $f_l(\ub)$ (HF function) subject to $K$ unknown constraints.\\
    \textbf{Define:} AFs (see Eq.\ref{eq: AF_C_LF}, Eq.\ref{eq: AF_HF_c}, and Eq.\ref{eq: composite_AF}) and $\epsilon$ , $v$ (see Eq.\ref{eq: threshold}).\\
    \While{stop conditions not met}{
        \begin{enumerate}
            \item Train a MF GP on objective function values using $\{\ub^{(i)}, y_j(\ub^{(i)})\}_{i=1}^q$ data.
            \item Train $K$ MF GPs on constraint values using $\{\ub^{(i)}, g_j^{k}(\ub^{(i)})\}_{i=1}^q$, $k=1,\dots,K$ data.
            \item Calculate the constrained cost-aware MF acquisition values:
            \[
            \gamma_{\cmfbo}(\ub; j) = 
            \begin{cases}
                \frac{\gamma_{C-LF}(\ub; j)}{O(j)} & \text{for } j = 1, \dots, ds \text{ and } j \neq l \\
                \frac{\gamma_{C-HF}(\ub; l)}{O(l)} & \text{for } j = l
            \end{cases}
            \]
            \item Solve the auxiliary optimization problem: 
            \[
            [\ub^{(q+1)}, j^{(q+1)}] = \underset{\ub, j}{\operatorname{argmax}} \, \gamma_{\cmfbo}(\ub ; j)
            \]
            \item Query $f_j(\cdot)$ and $G_j^{k}(\cdot)$ at point $\ub^{(q+1)}$ from source $j^{(q+1)}$ to obtain $y_{j^{(q+1)}}(\ub^{(q+1)})$ and $g_{j^{(q+1)}}^{k}(\ub^{(q+1)}), \forall k$.
            \item Update data: 
            \[
            \mathcal{D}^{q+1} \leftarrow \mathcal{D}^{q} \cup (\ub^{(q+1)}, y_{j^{(q+1)}}(\ub^{(q+1)}), g_{j^{(q+1)}}^{k}(\ub^{(q+1)}))
            \]
            \item Perform post-acquisition optimization to find ${y}_{POA}^{* (q)}$ and normalize them.
            \item Check the convergence criteria:
            \[var(\{\tilde{y}_{POA}^{*(i)}\}_{i=q-v+1}^{q}) < \epsilon\]
            \item Update counter: $q \leftarrow q+1$
        \end{enumerate}       
    }
    \textbf{Output:} Updated data $\mathcal{D}^{q}=\{(\ub^{(i)}, y_j(\ub^{(i)}), g_j^{k}(\ub^{(i)}))\}_{i=1}^q$, GPs trained on objective and constraint functions.    \caption{Constrained Cost-aware MFBO}
    \label{alg: MFBO-algorithm}
\end{algorithm*}

    \section{Results and Discussions} \label{sec: results}
In this section, we demonstrate the performance of \cmfbo~on five benchmark problems whose input dimensionality ranges from $3$ to $20$. Detailed descriptions of the number of initial data points, sampling costs, the relative accuracy of LF to HF data (which is not used by \cmfbo), and the added noise are provided in Table\ref{table: analytic-formulation}. For SF cases, we adjust the number of initial HF samples to match the initial cost of their MF counterparts. The functional forms are provided in \cite{gp_plus}.
In our studies, we assume that the cost of querying any data source significantly outweighs the computational costs of BO (e.g., fitting emulators or solving auxiliary optimization problems). Consequently, the primary metric for comparison is each method’s ability to identify the global optimum of the HF source while minimizing overall data collection costs.
\begin{table*}[b!]
\centering
\scriptsize
\begin{tabular}{cccccccccccccccccc}
\hline
\multirow{2}{*}{\begin{tabular}[c]{@{}c@{}}Function\\ Name\end{tabular}} & \multicolumn{5}{c}{n}      & Small Noise & Large Noise & \multicolumn{4}{c}{RRMSE} & \multicolumn{5}{c}{Cost}     & \multirow{2}{*}{K} \\ \cline{2-17}
    & HF & LF1 & LF2 & LF3 & LF4 & HF          & HF          & LF1  & LF2  & LF3  & LF4  & HF   & LF1 & LF2 & LF3 & LF4 &                    \\ \hline
\braninh                                                               & 9  & 18  & -   & -   & -   & 0.1         & 0.2         & 2.82 & -    & -    & -    & 10   & 1   & -   & -   & -   & 2                  \\
\hartman                                                                  & 18 & 36  & -   & -   & -   & 0.25        & 0.5         & 0.79 & -    & -    & -    & 10   & 1   & -   & -   & -   & 2                  \\
\wing                                                                     & 30 & 60  & 60  & 60  & -   & 0.5         & 1           & 0.19 & 1.14 & 5.74 & -    & 1000 & 100 & 10  & 1   & -   & 1                  \\
\sepwing                                                                & 30 & 60  & 60  & 60  & -   & 0.5         & 1           & 0.19 & 1.14 & 5.74 & -    & 1000 & 100 & 10  & 1   &     & 1                  \\
\twenty                                                                     & 30 & 60  & 60  & 60  & 60  & 0.5         & 1           & 0.23 & 0.48 & 0.23 & 0.25 & 200  & 100 & 50  & 10  & 5   & 1                  \\ \hline
\end{tabular}
\caption{\textbf{Analytic Functions}: $K$ denotes the number of constraints and RRMSE of an LF source is calculated by comparing its output to that of the HF source at $10000$ random points. The cost column lists the cost of obtaining a sample from the corresponding source, while the small/large noise indicates the standard deviation of the added noise. In all examples expect \wing, the constraints vary across the sources.}
    \label{table: analytic-formulation}
\end{table*}
\subsection{Benchmark Setup}
To evaluate the effectiveness of \cmfbo, we compare it against two popular SF baselines from \texttt{botorch}, namely, Turbo and qEI. To further illustrate the versatility of our approach, we also include an SF version of \cmfbo, referred to as \csfbo.
For Turbo and qEI, we use a fixed number of iterations ($60$) to stop BO while for \cmfbo~and \csfbo~we utilize the stopping criterion described in Section\ref{sec: stop} with parameters $v=10$ and $\epsilon=0.01$ where the latter parameter is increased to $\epsilon=0.5$ for the $20D$ example.
To challenge the convergence and robustness of \cmfbo, we introduce noise exclusively to the HF data. The noise variance is determined based on the range of each function, and each problem is tested under small and large noise variances.
Lastly, to quantify the impact of random initial data, we repeat the optimization $10$ times for each example and each method. The initial data for all runs are generated using the Sobol sequence.

\subsection{Summary of Results}
\begin{figure*}[!h]
    \centering
    \begin{subfigure}{.48\textwidth}
        \centering
        \includegraphics[width=0.89\linewidth]{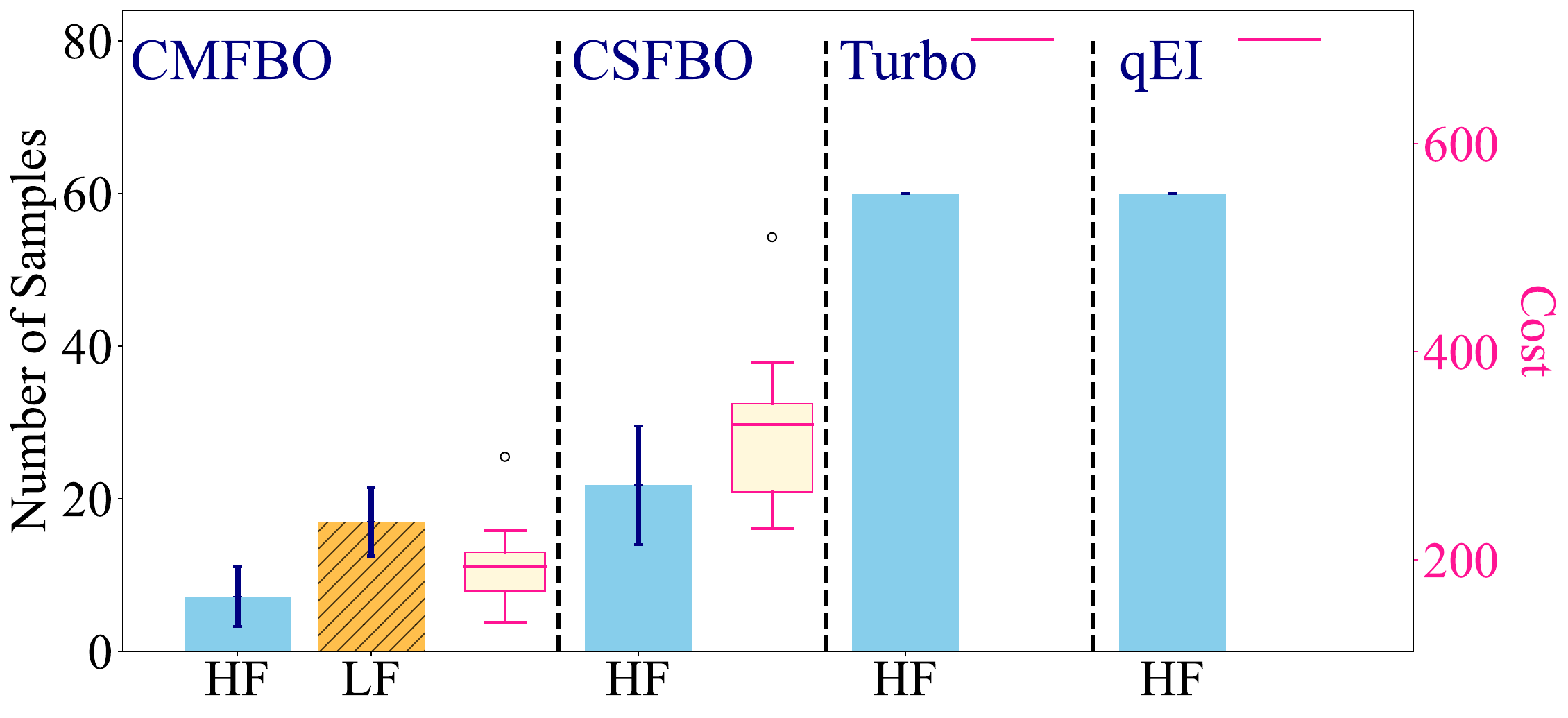}
        \caption{\braninh~($3D$)}
        \label{fig: cost_case3_01}
    \end{subfigure}%
    \begin{subfigure}{.47\textwidth}
        \centering
        \includegraphics[width=0.89\linewidth]{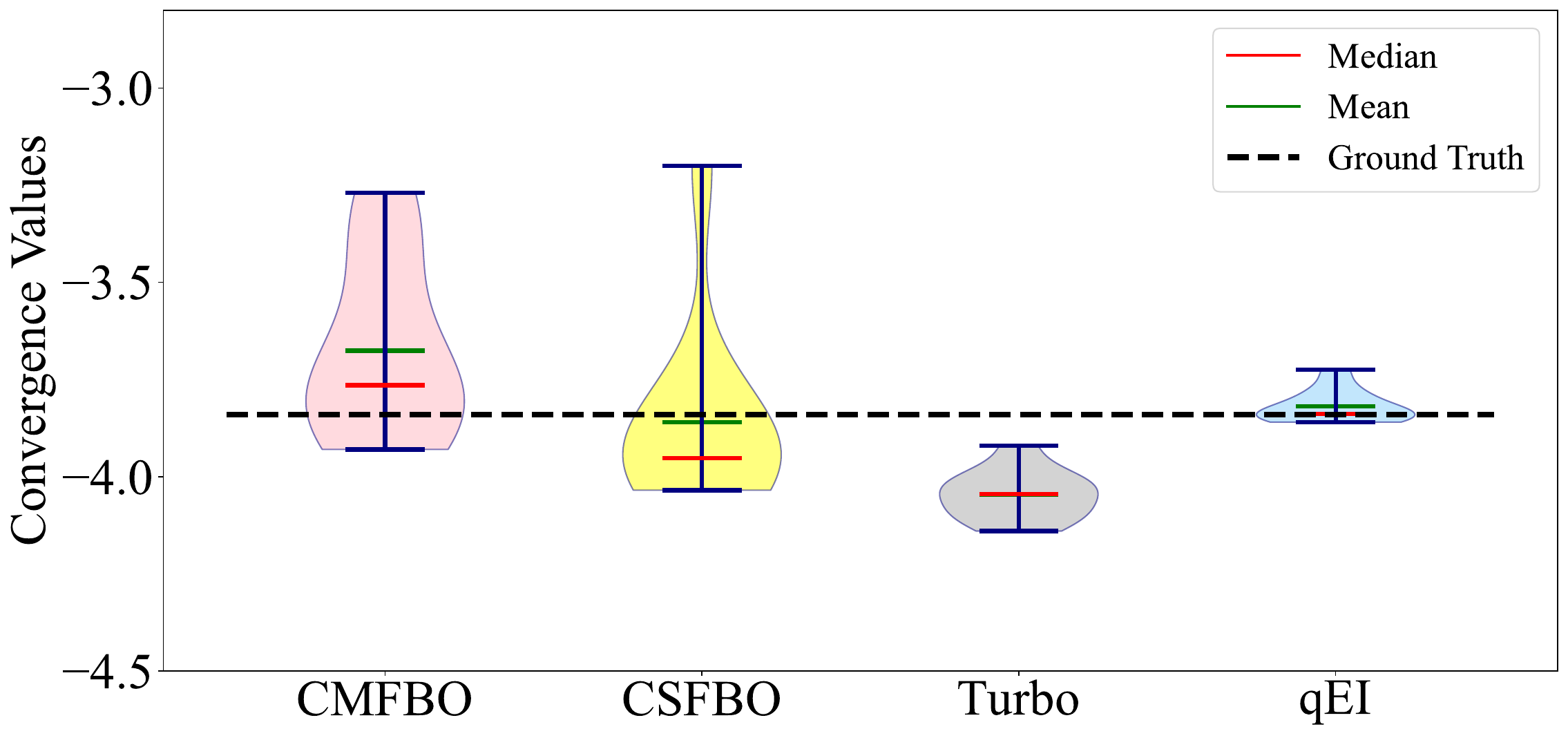}
        \caption{\braninh~($3D$)}
        \label{fig: converge_case3_01}
    \end{subfigure}
    \newline
    \begin{subfigure}{0.48\textwidth}
        \centering
        \includegraphics[width=0.89\linewidth]{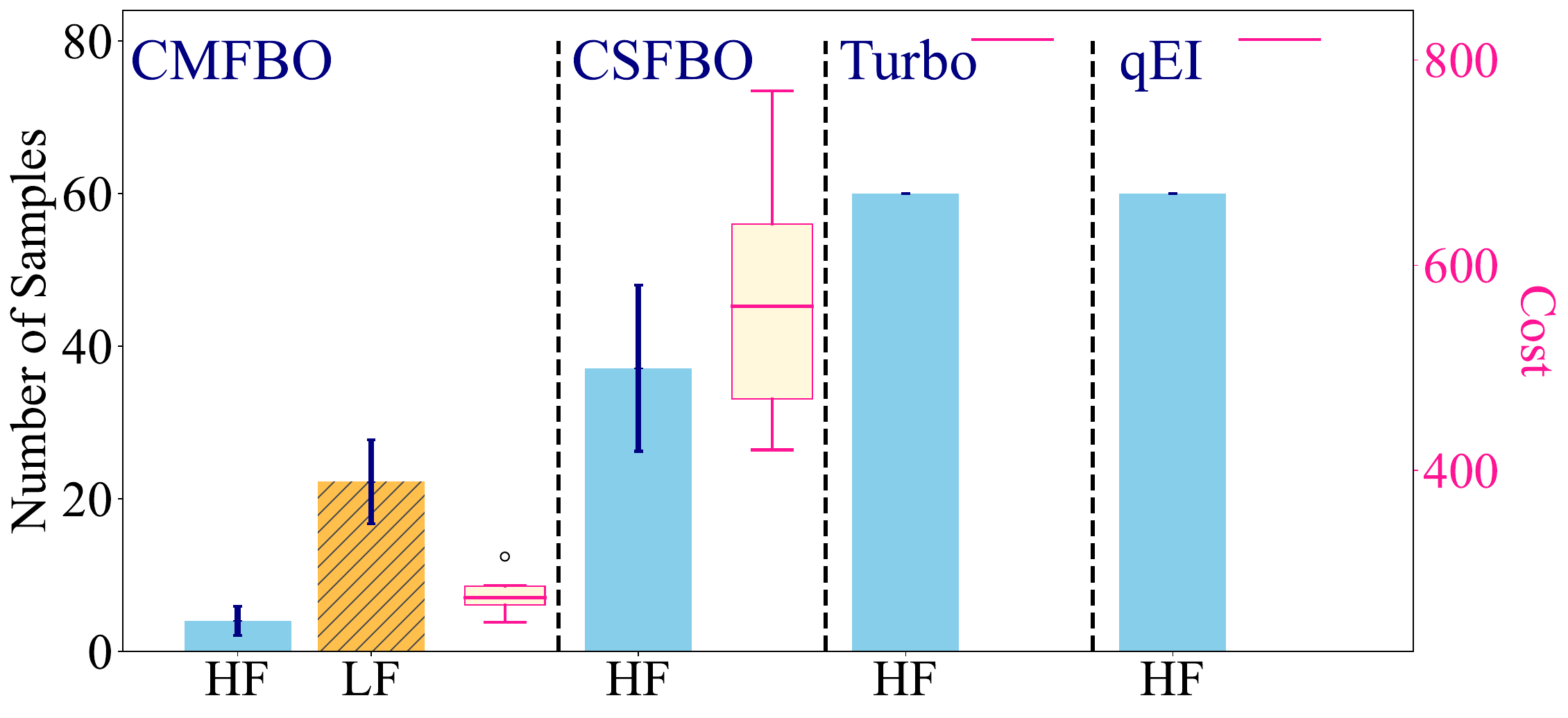}
        \caption{\hartman~($6D$)}
        \label{fig: cost_Hartman_0.25}
    \end{subfigure}%
    \begin{subfigure}{.47\textwidth}
        \centering
        \includegraphics[width=0.89\linewidth]{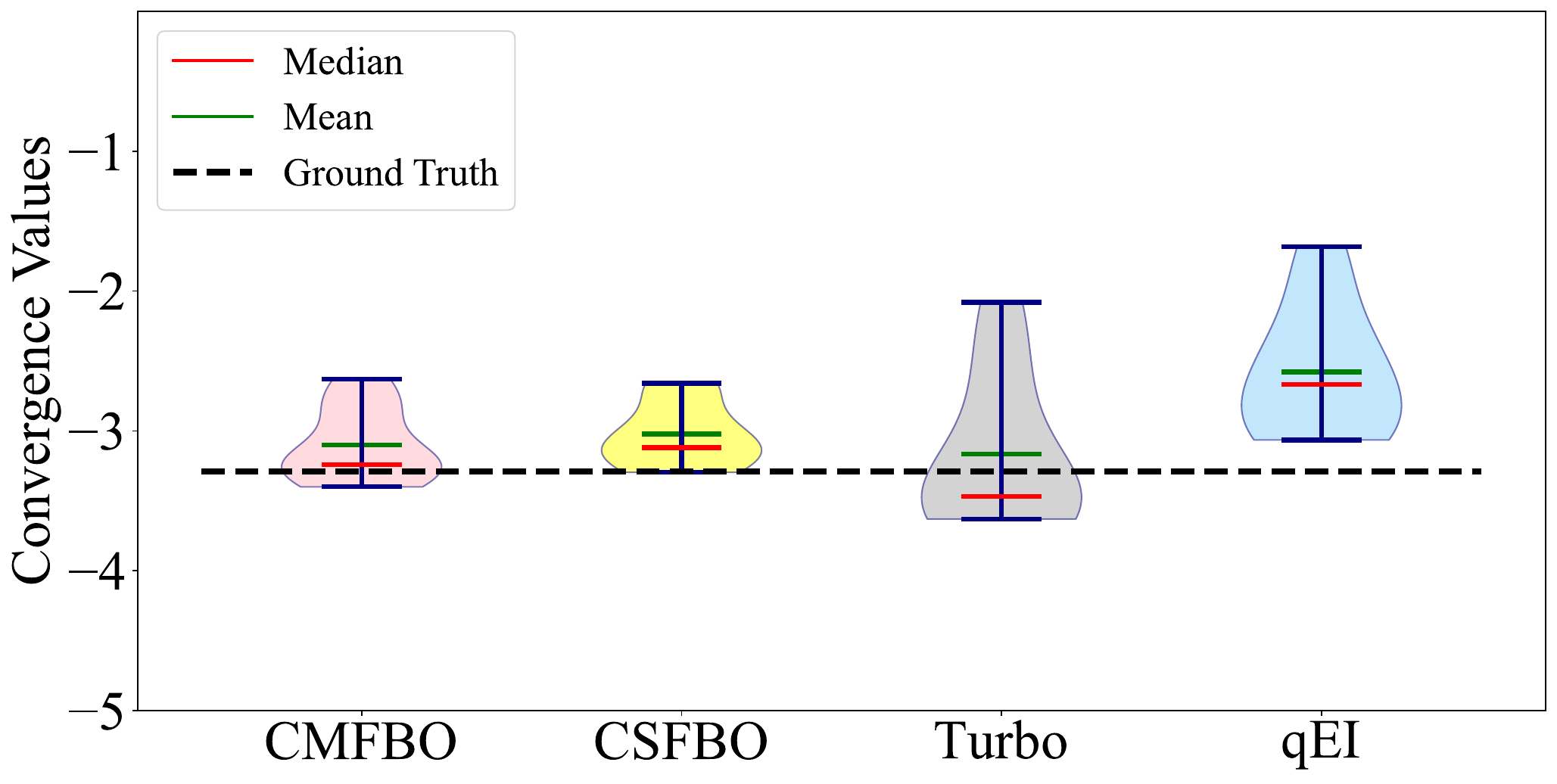}
        \caption{\hartman~($6D$)}
        \label{fig: converge_Hartman_0.25}
    \end{subfigure}
    \newline
    \begin{subfigure}{0.48\textwidth}
        \centering
        \includegraphics[width=0.89\linewidth]{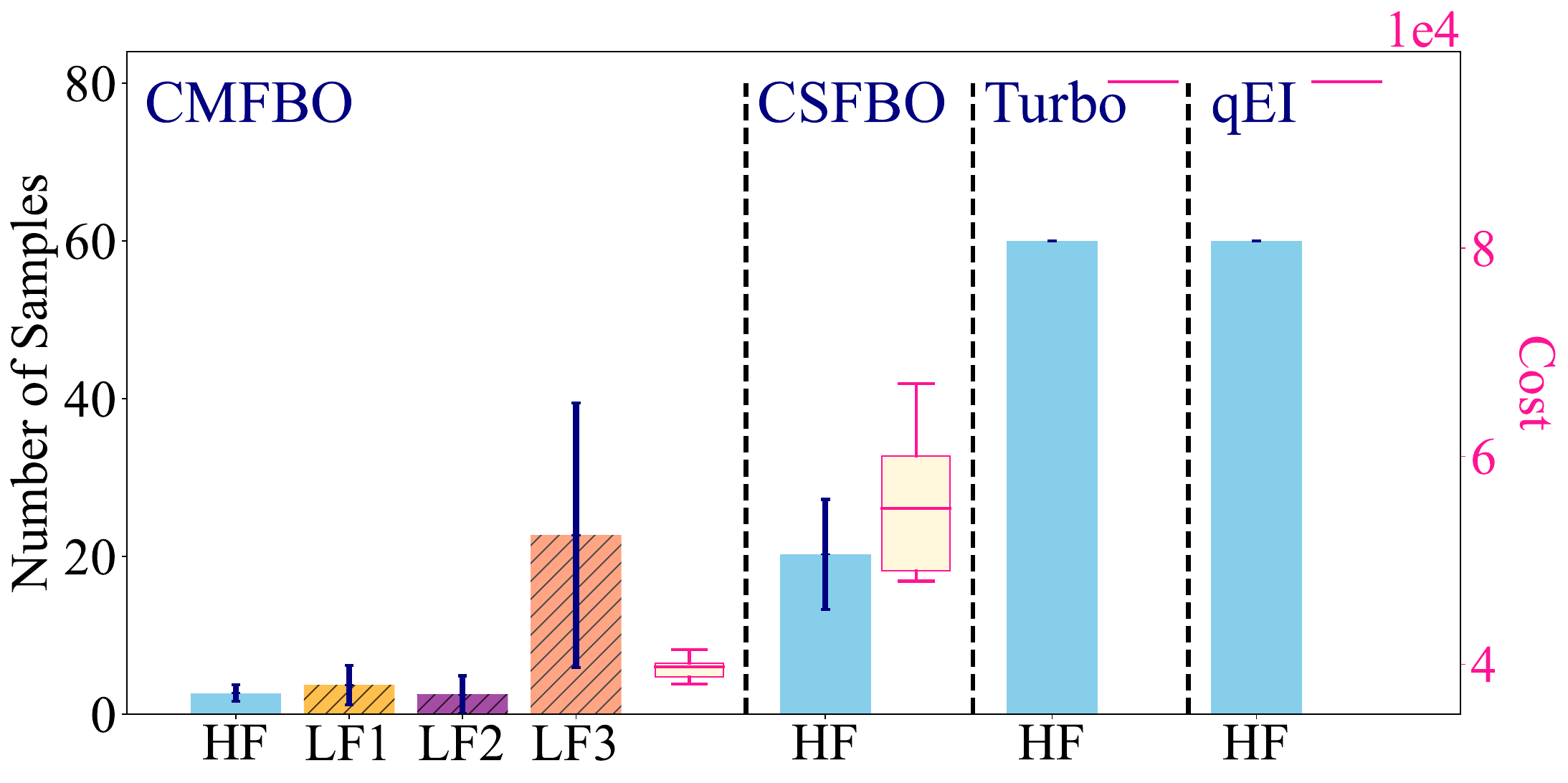}
        \caption{\wing~($10D$)}
        \label{fig: cost_wing_0.5}
    \end{subfigure}%
    \begin{subfigure}{.48\textwidth}
        \centering
        \includegraphics[width=0.89\linewidth]{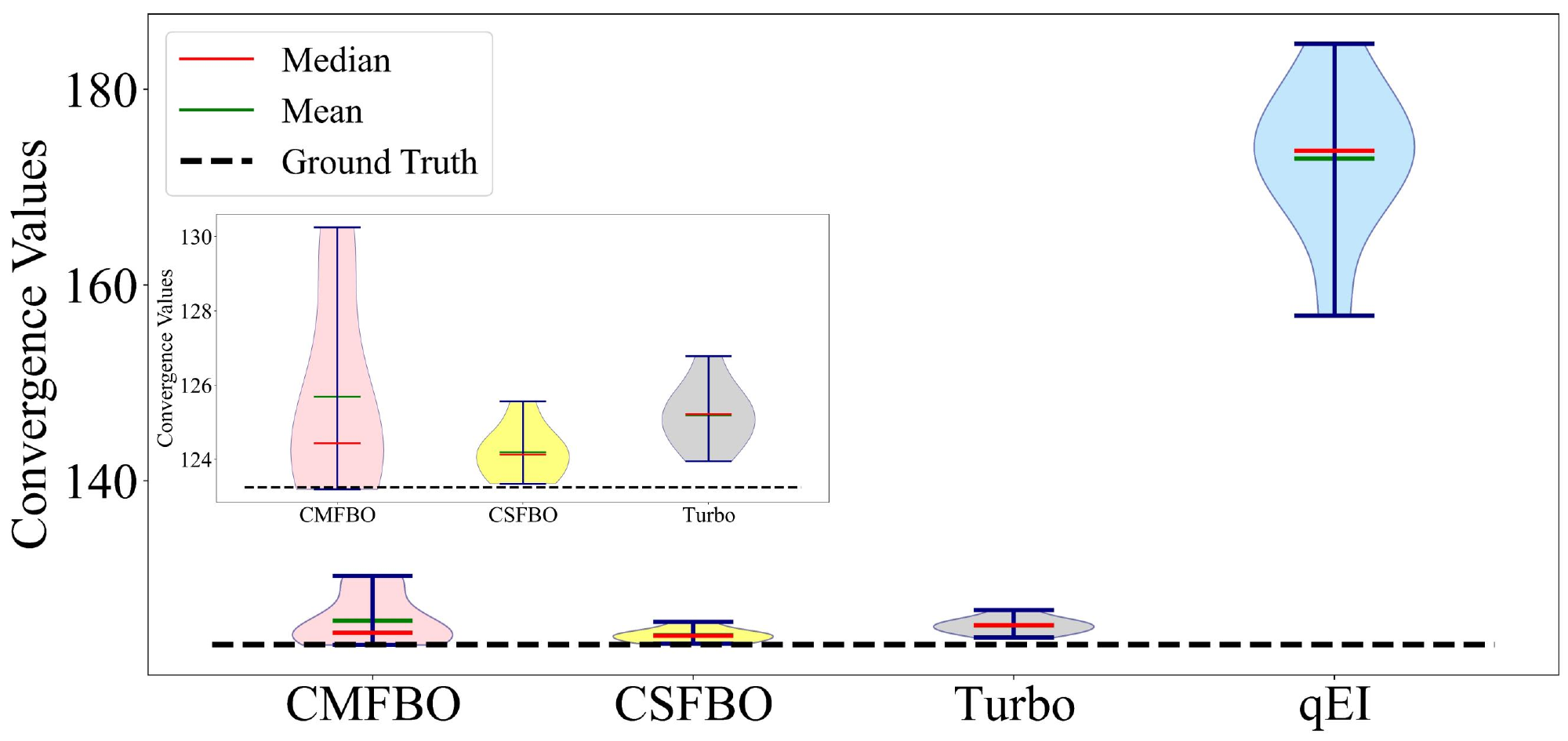}
        \caption{\wing~($10D$)}
        \label{fig: converge_wing_0.5}
    \end{subfigure}
    \newline
    \begin{subfigure}{0.48\textwidth}
        \centering
        \includegraphics[width=0.89\linewidth]{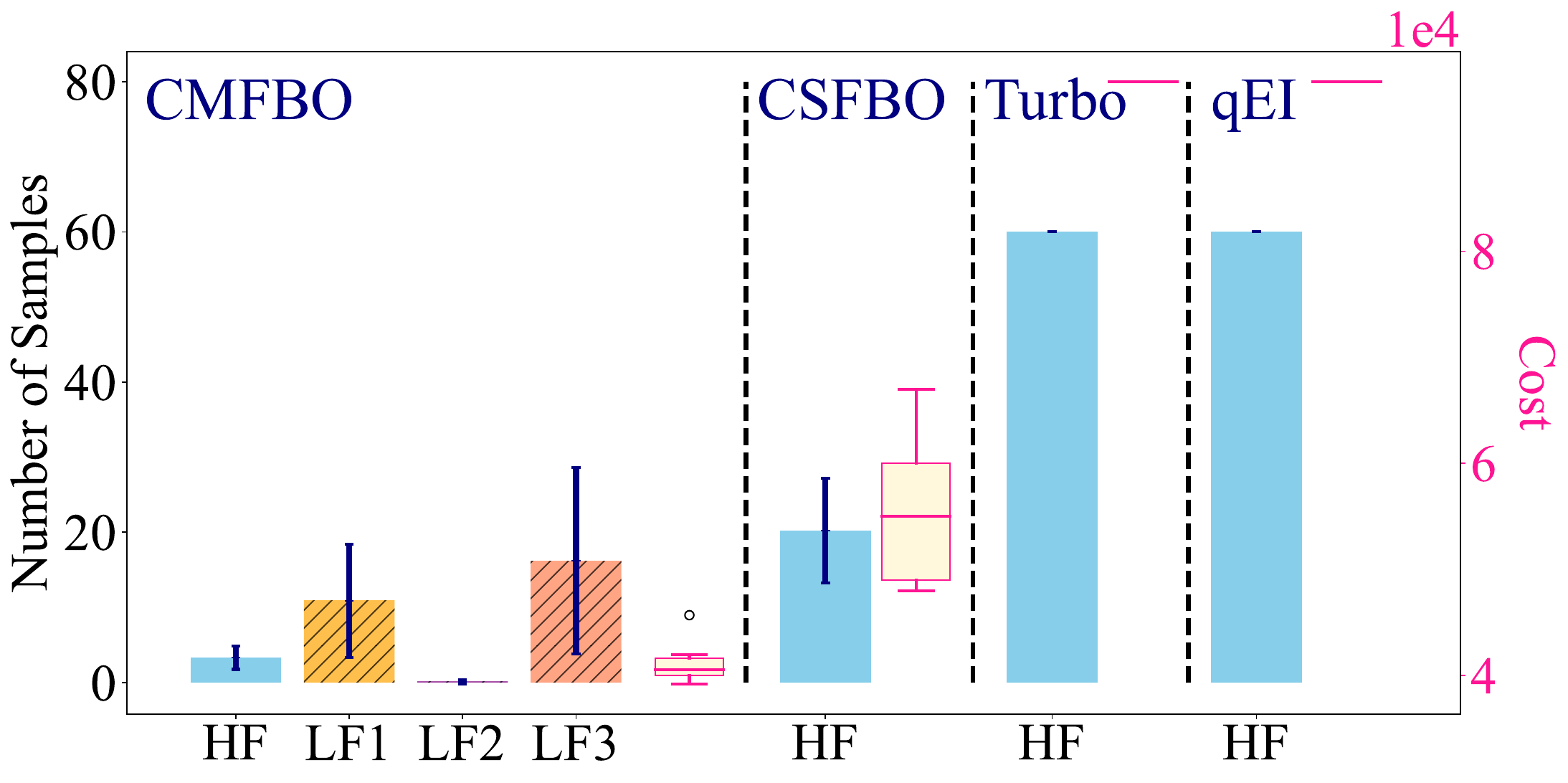}
        \caption{\sepwing~($10D$)}
        \label{fig: cost_wing_0.5_sep}
    \end{subfigure}%
    \begin{subfigure}{.48\textwidth}
        \centering
        \includegraphics[width=0.89\linewidth]{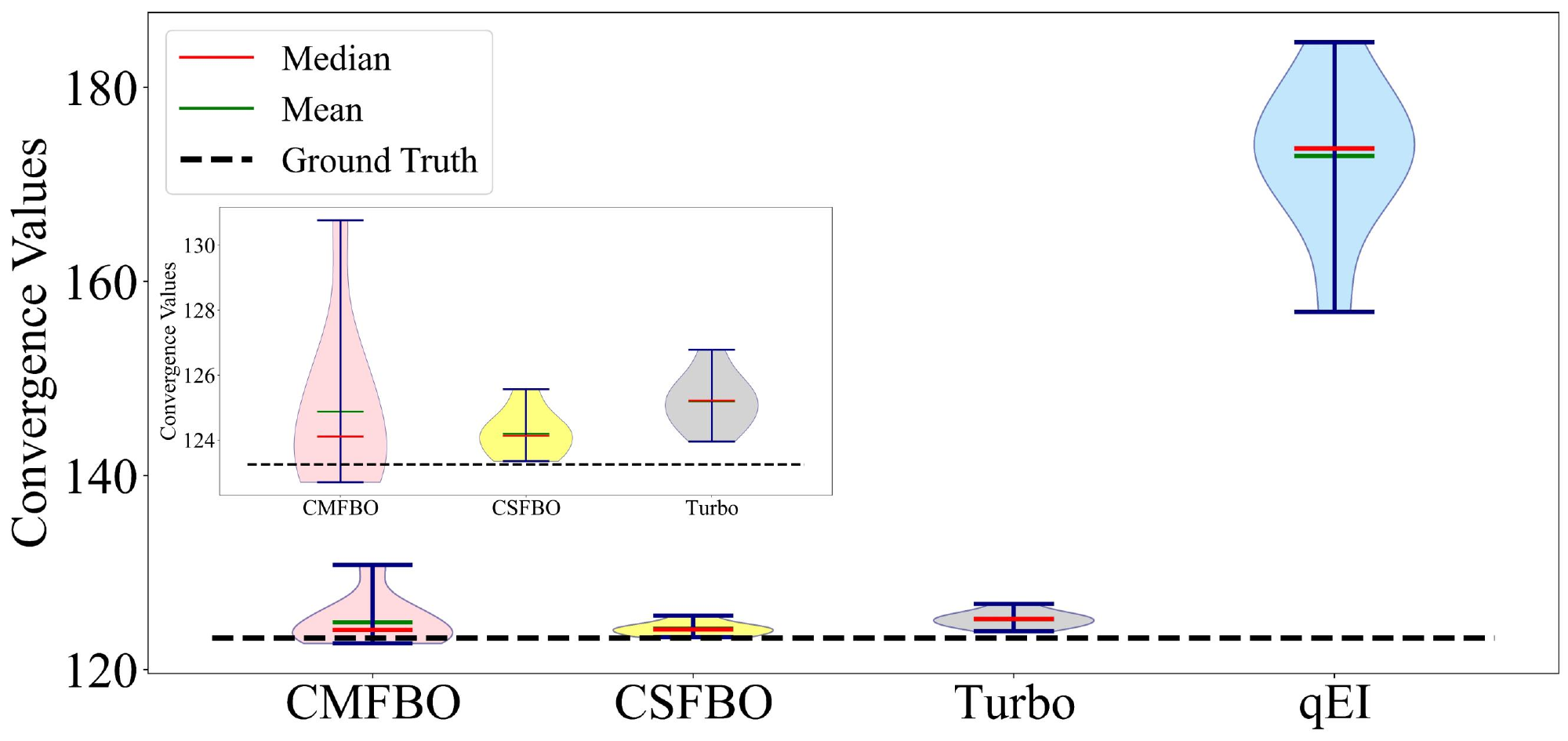}
        \caption{\sepwing~($10D$)}
        \label{fig: converge_wing_0.5_sep}
    \end{subfigure}
    \newline
    \begin{subfigure}{0.48\textwidth}
        \centering
        \includegraphics[width=0.89\linewidth]{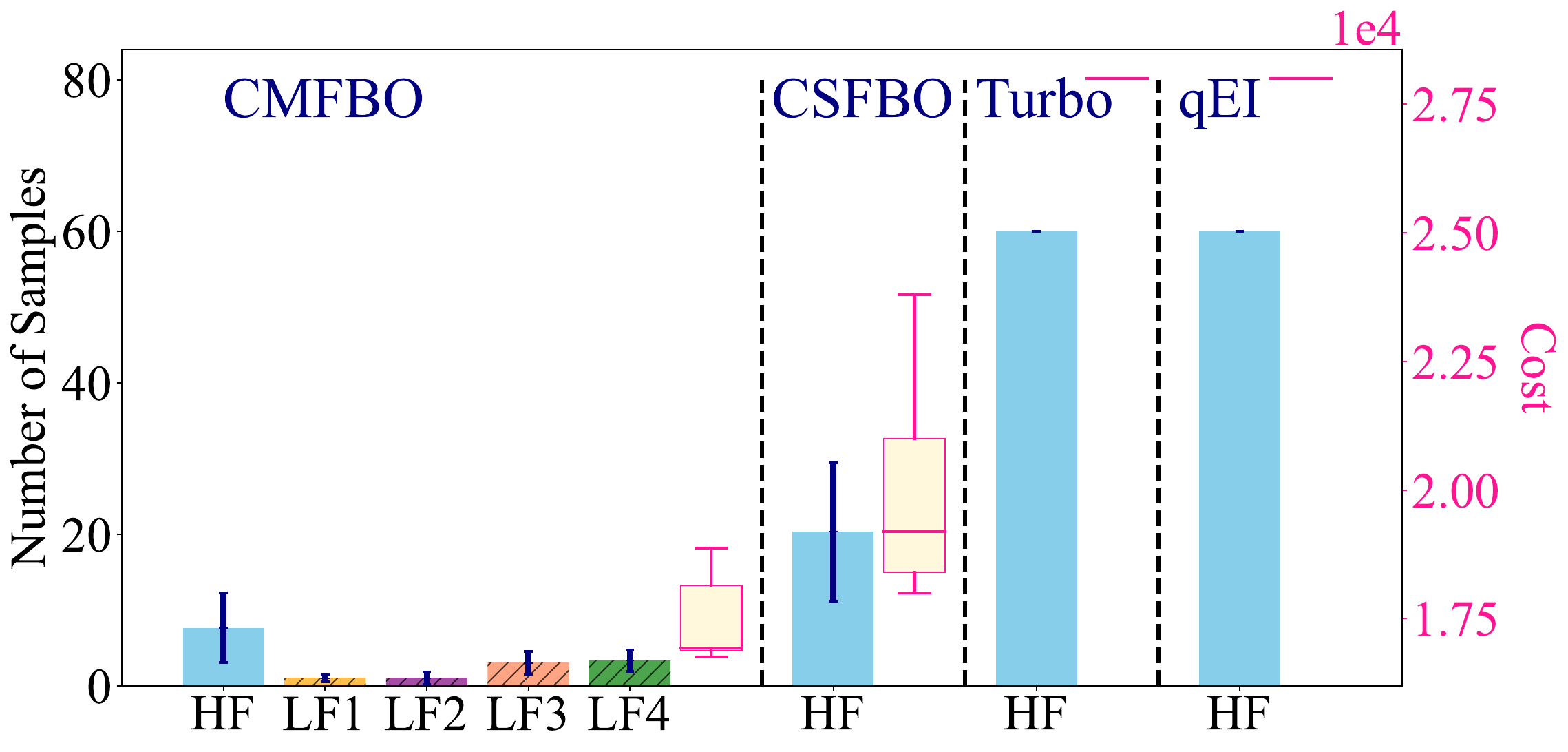}
        \caption{\twenty~($20D$)}
        \label{fig: cost_20d_0.5_sep}
    \end{subfigure}%
    \begin{subfigure}{.48\textwidth}
        \centering
        \includegraphics[width=0.89\linewidth]{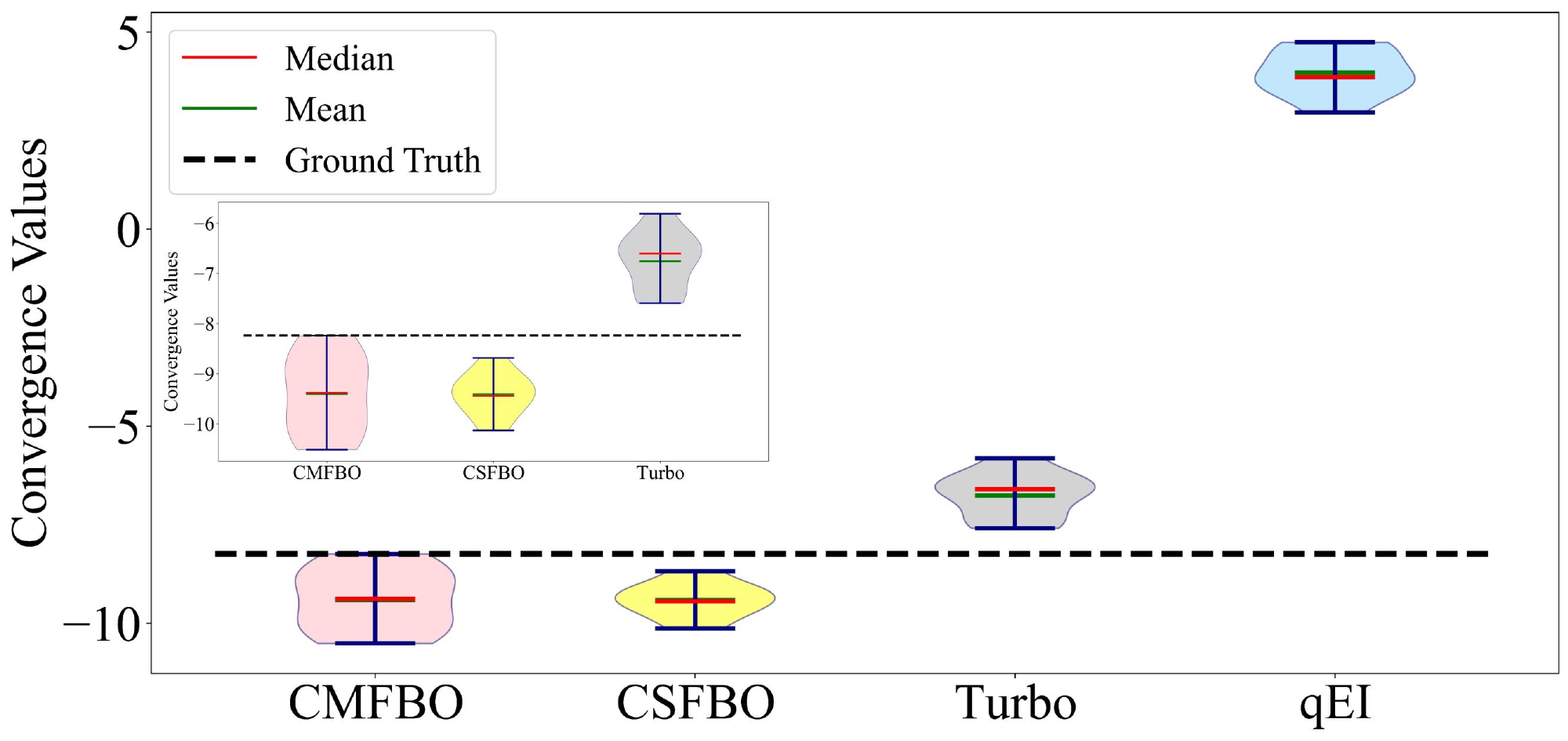}
        \caption{\twenty~($20D$)}
        \label{fig: converge_20d_0.5_sep}
    \end{subfigure}
    \caption{\textbf{Convergence values and costs (small noise):} The insets provide a magnified view of the variations.}
    \label{fig: plots_small_noise}
\end{figure*}

\begin{figure*}[!ht]
    \centering
    \begin{subfigure}{.47\textwidth}
        \centering
        \includegraphics[width=0.95\linewidth]{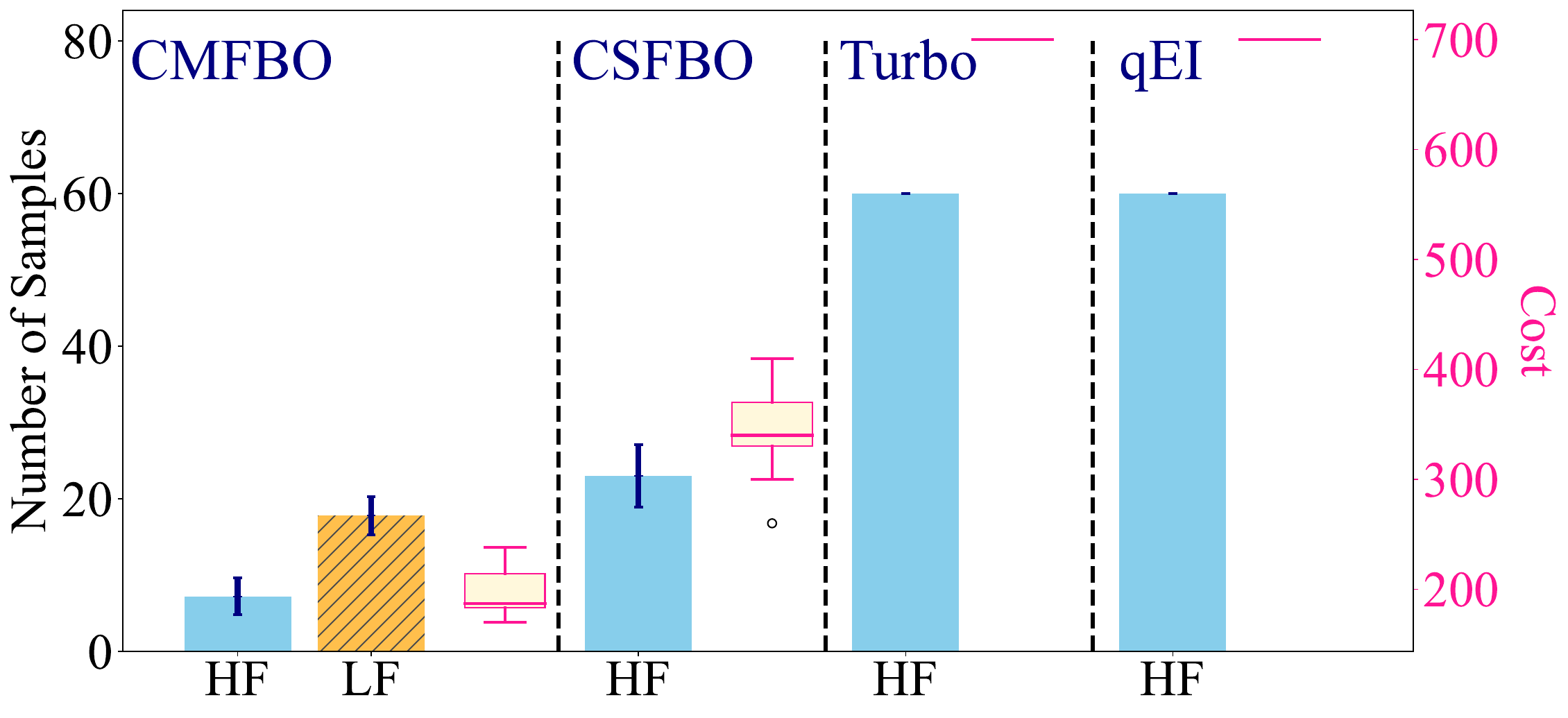}
        \caption{\braninh~($3D$)}
        \label{fig: cost_case3_02}
    \end{subfigure}%
    \begin{subfigure}{.47\textwidth}
        \centering
        \includegraphics[width=0.9\linewidth]{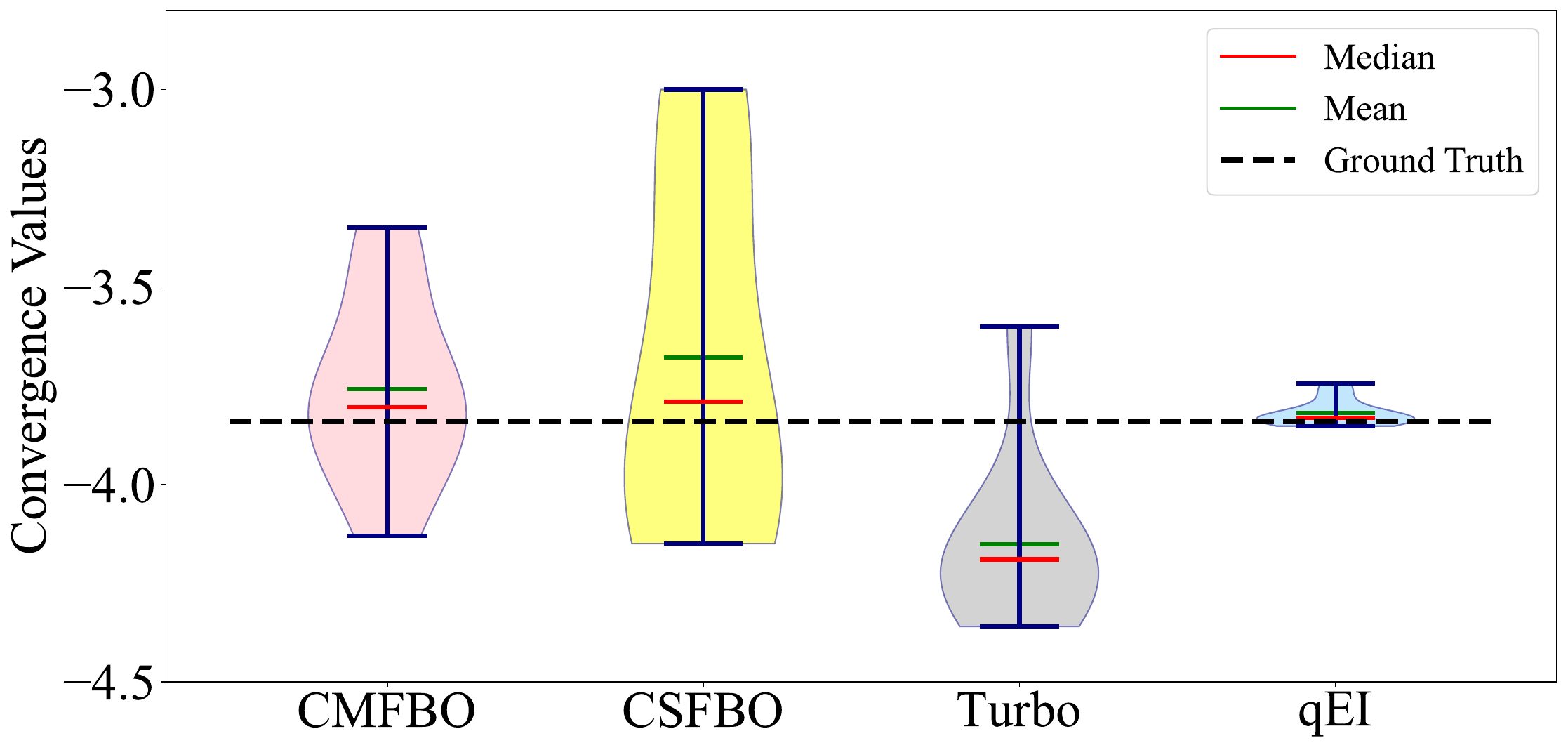}
        \caption{\braninh~($3D$)}
        \label{fig: converge_case3_02}
    \end{subfigure}
    \newline
    \begin{subfigure}{0.47\textwidth}
        \centering
        \includegraphics[width=0.95\linewidth]{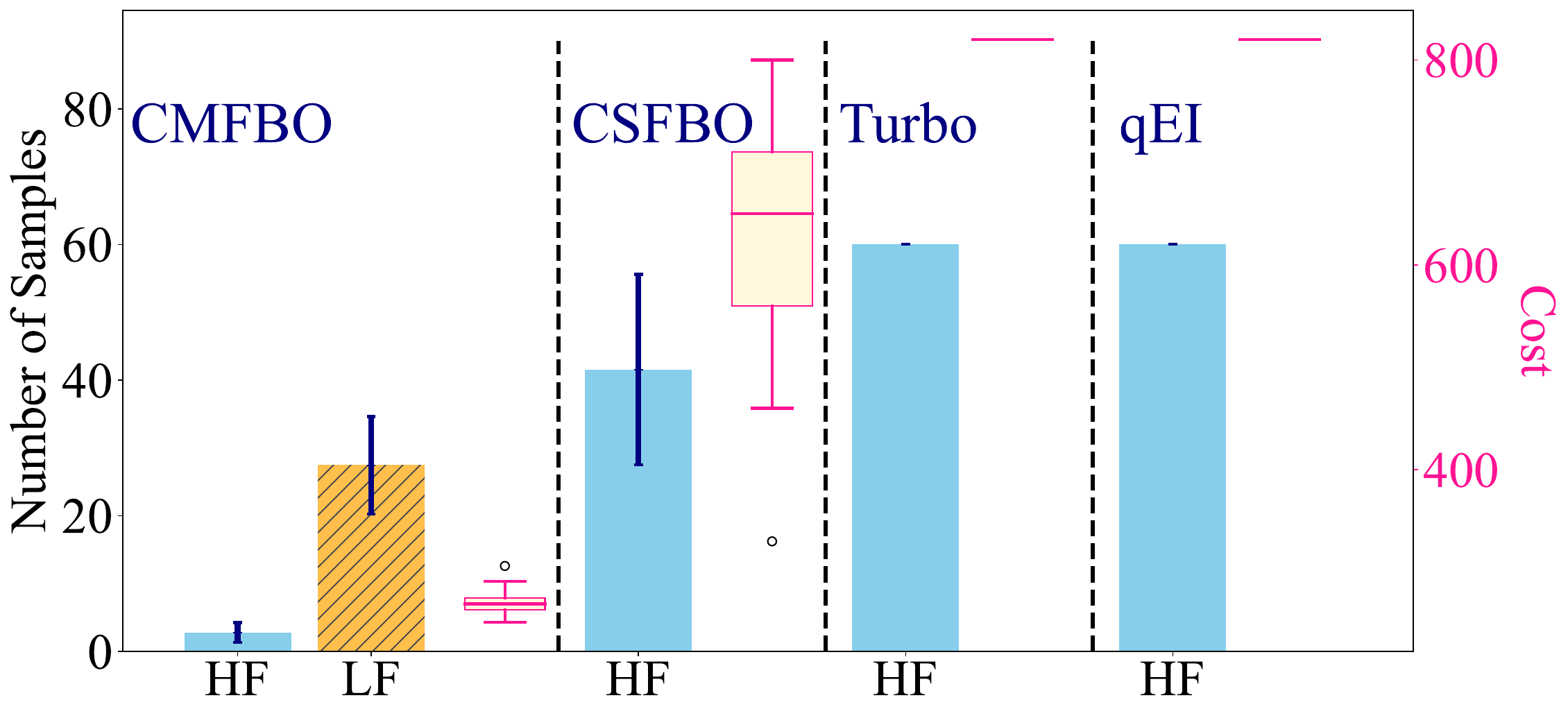}
        \caption{\hartman~($6D$)}
        \label{fig: cost_Hartman_0.5}
    \end{subfigure}%
    \begin{subfigure}{.47\textwidth}
        \centering
        \includegraphics[width=0.89\linewidth]{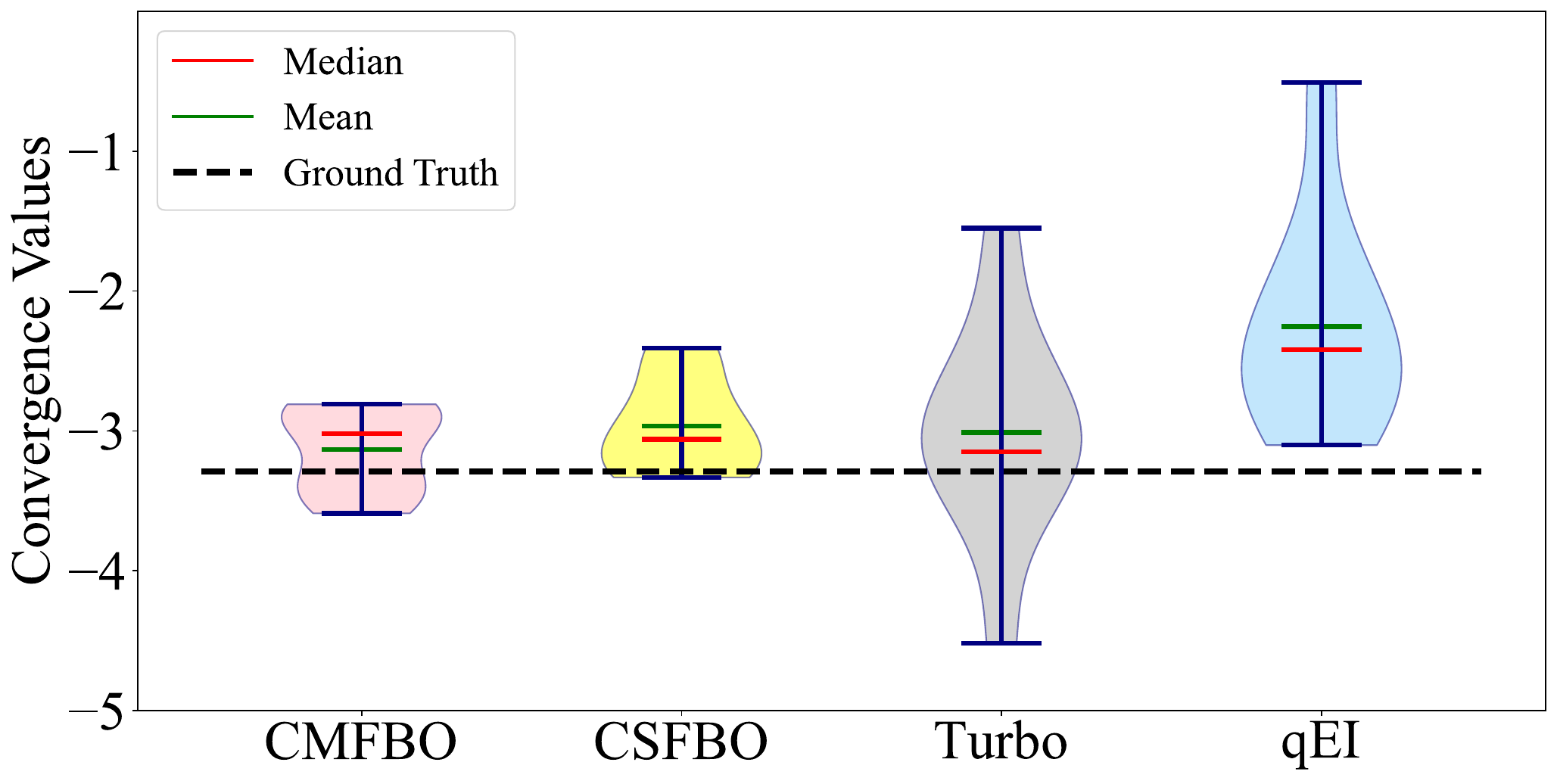}
        \caption{\hartman~($6D$)}
        \label{fig: converge_Hartman_0.5}
    \end{subfigure}
    \newline
    \begin{subfigure}{0.47\textwidth}
        \centering
        \includegraphics[width=0.9\linewidth]{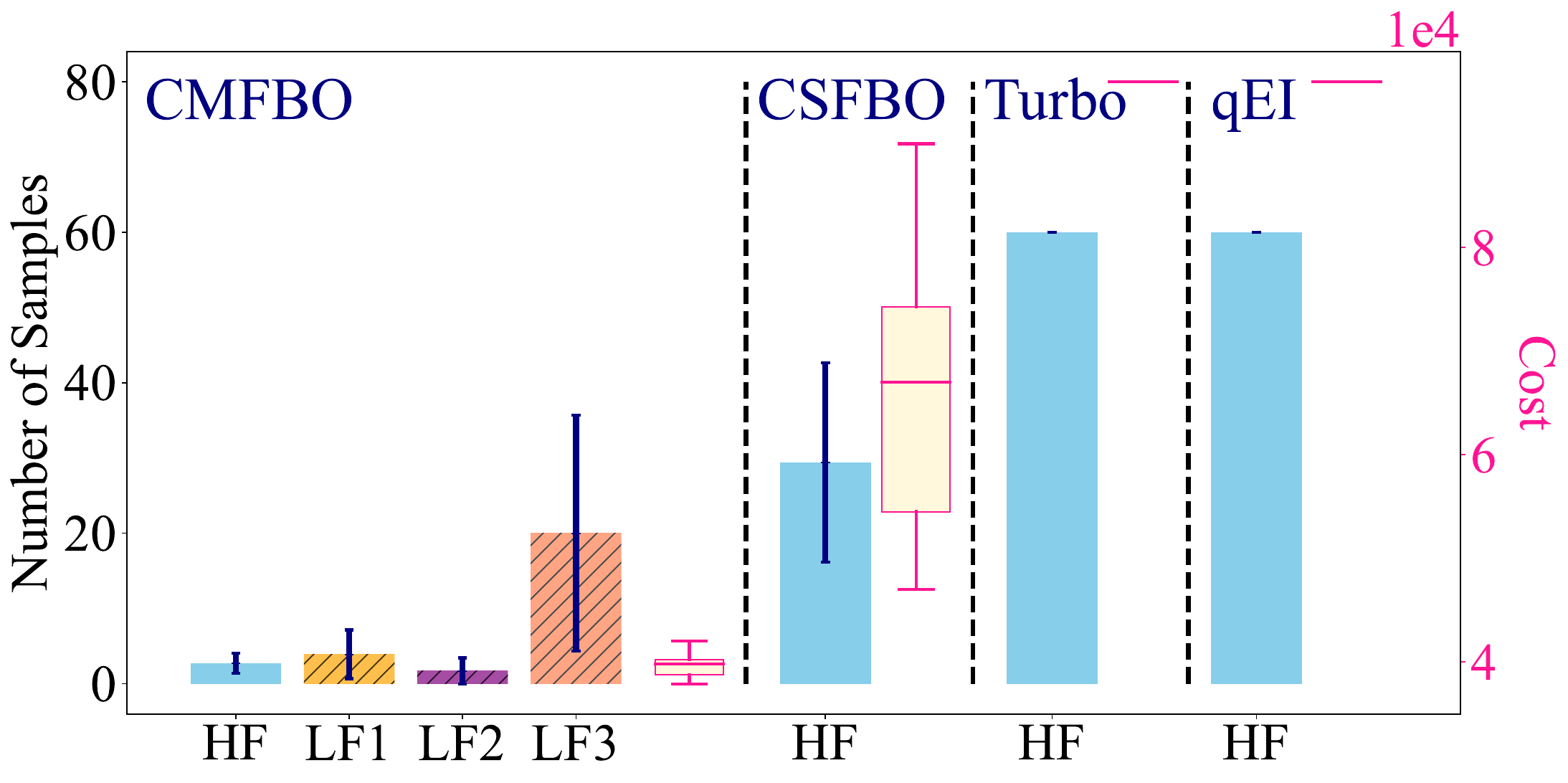}
        \caption{\wing~($10D$)}
        \label{fig: cost_wing_1}
    \end{subfigure}%
    \begin{subfigure}{.47\textwidth}
        \centering
        \includegraphics[width=0.9\linewidth]{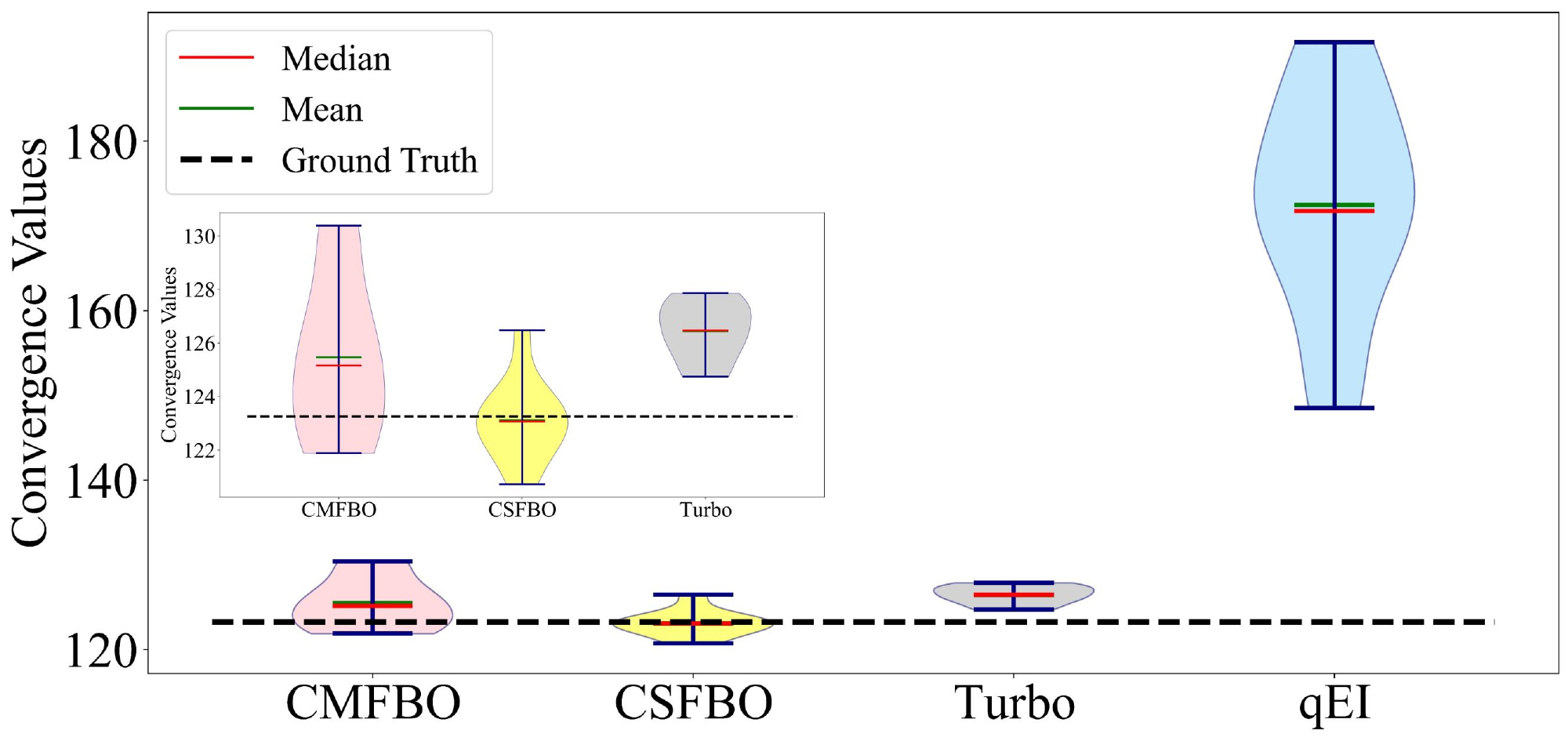}
        \caption{\wing~($10D$)}
        \label{fig: converge_wing_1}
    \end{subfigure}
    \newline
    \begin{subfigure}{0.47\textwidth}
        \centering
        \includegraphics[width=0.9\linewidth]{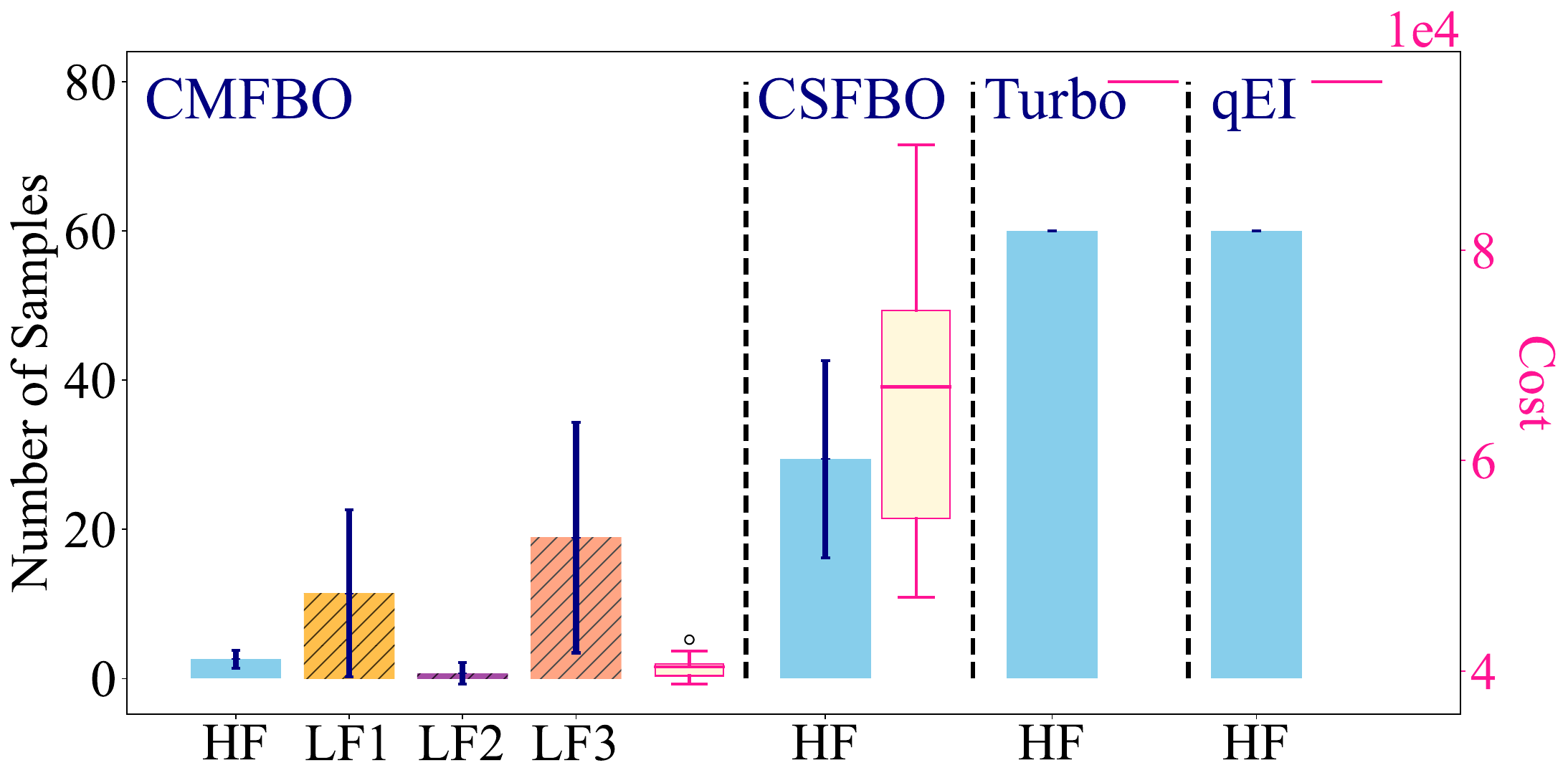}
        \caption{\sepwing~($10D$)}
        \label{fig: cost_wing_1_sep}
    \end{subfigure}%
    \begin{subfigure}{.47\textwidth}
        \centering
        \includegraphics[width=0.9\linewidth]{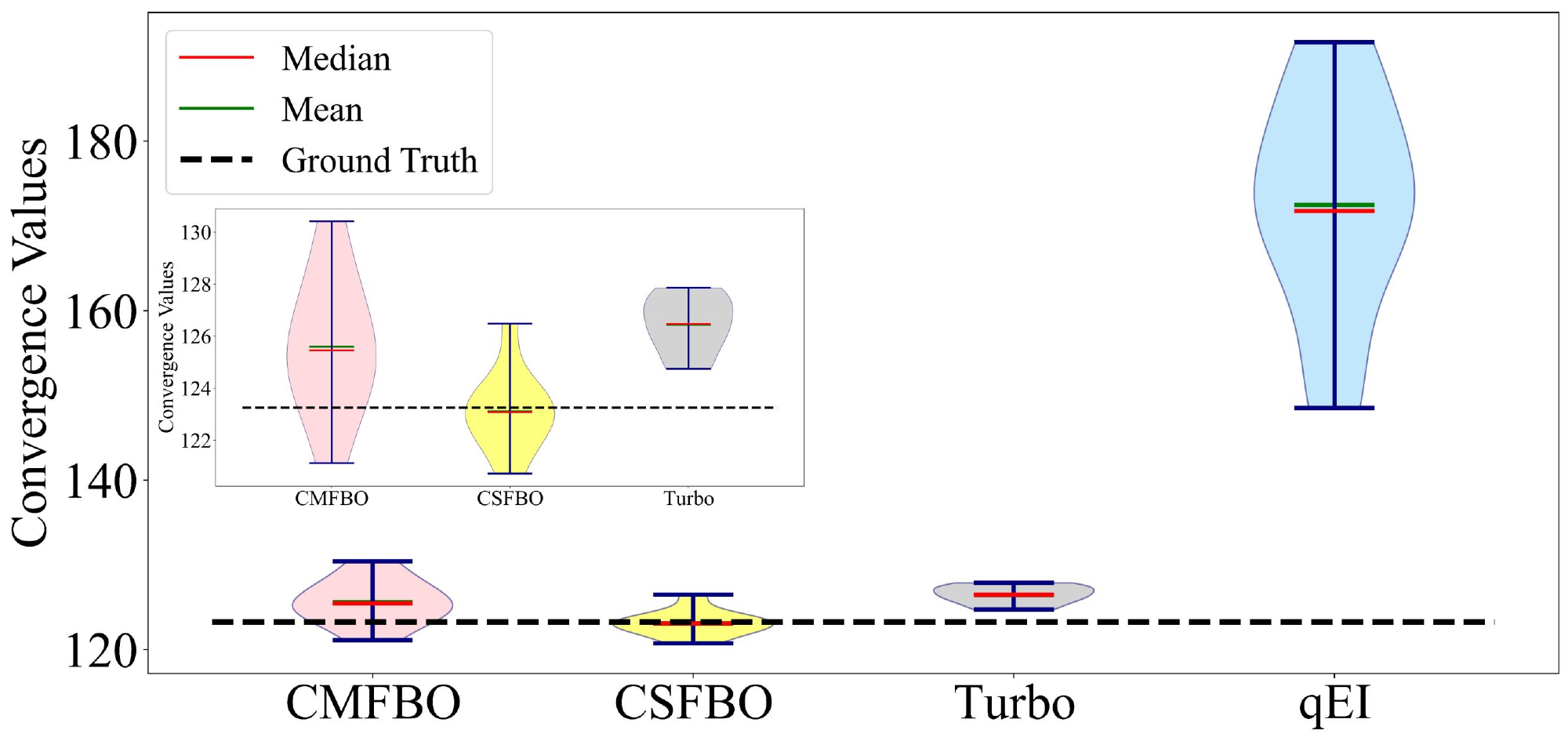}
        \caption{\sepwing~($10D$)}
        \label{fig: converge_wing_1_sep}
    \end{subfigure}
    \newline
    \begin{subfigure}{0.47\textwidth}
        \centering
        \includegraphics[width=0.9\linewidth]{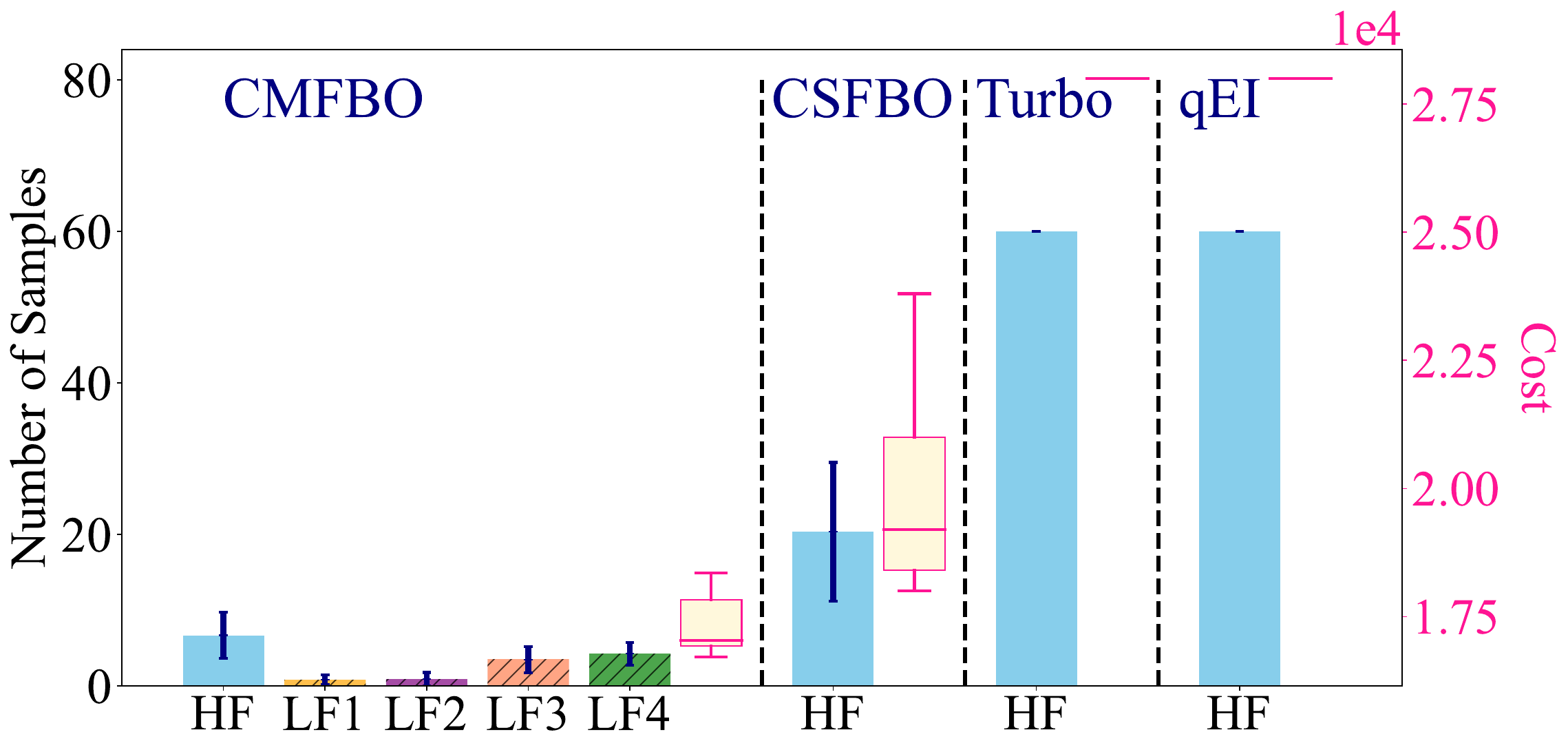}
        \caption{\twenty~($20D$)}
        \label{fig: cost_20d_1_sep}
    \end{subfigure}%
    \begin{subfigure}{.47\textwidth}
        \centering
        \includegraphics[width=0.9\linewidth]{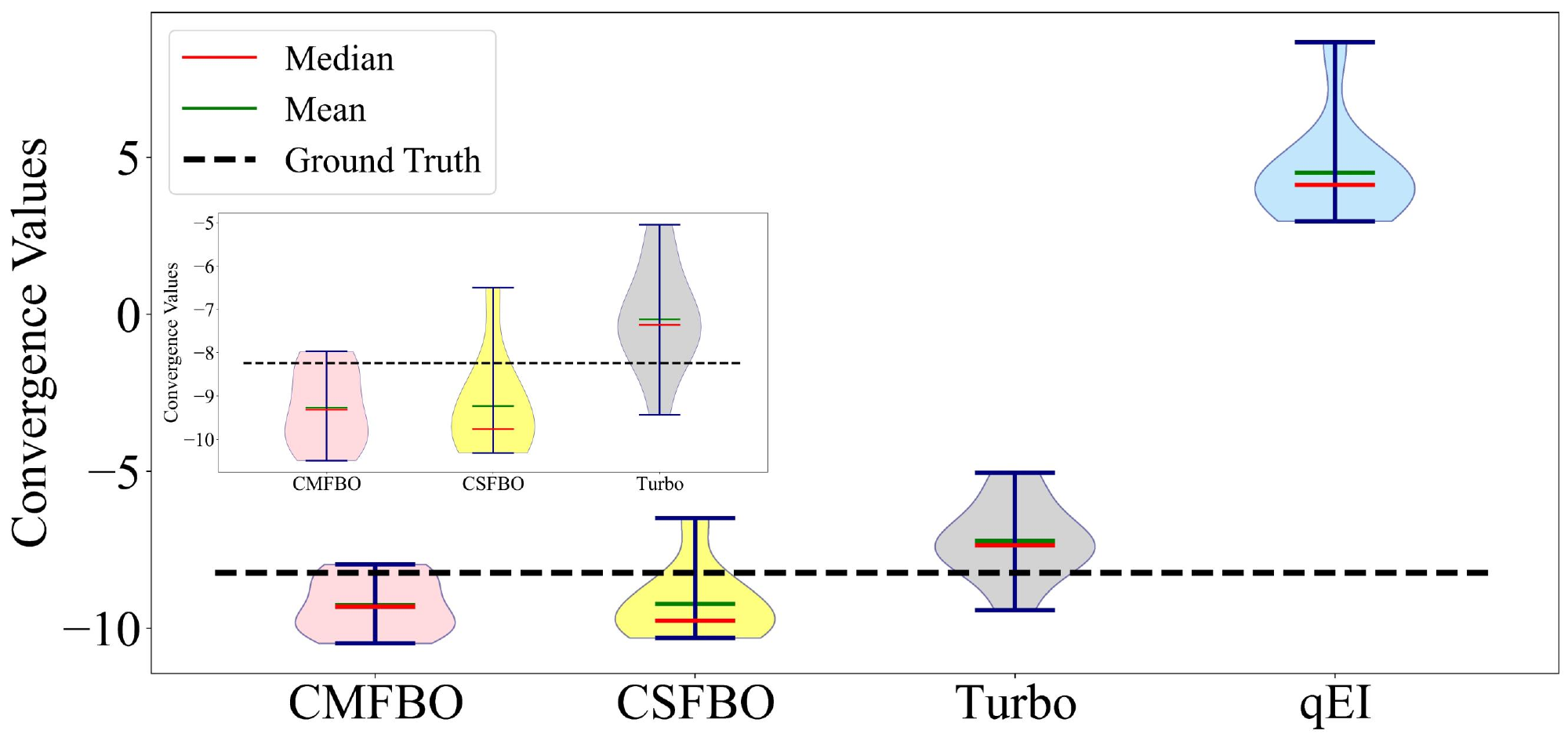}
        \caption{\twenty~($20D$)}
        \label{fig: converge_20d_1_sep}
    \end{subfigure}
    \caption{\textbf{Convergence values and costs (large noise):} The insets provide a magnified view of the variations.}
    \label{fig: plots_large_noise}
\end{figure*}

The results of our studies are illustrated in Fig.\ref{fig: plots_small_noise} and Fig.\ref{fig: plots_large_noise} where we observe that qEI exhibits strong robustness in low-dimensional problems but at a high convergence cost. However, its performance deteriorates as input dimensionality and noise variance increase: in $6D$ to $20D$ examples, qEI fails to converge to the ground truth even in a single repetition.
We also observe that Turbo, which is designed for scalable optimization, struggles to find an optimal solution in high-dimensional applications and its performance is sensitive to noise. 
Among the three SF methods, \csfbo~provides the lowest convergence cost and its performance, unlike qEI and Turbo, is quite robust to both added noise and increasing problem dimensionality. 

The proposed method in this paper, \cmfbo, provides the lowest sampling cost across all benchmark problems. \cmfbo~shows some sensitivity to the initial data and problem dimensionality which is due to the facts that it $(1)$ leverages far fewer HF samples than the other three methods, and $(2)$ LF sources provide biased values for either the objective function and/or constraints. 

\subsection{Discussions} 
Fig.\ref{fig: plots_small_noise} illustrates the performance of the four methods with small noise added to the HF samples. The left column of the figure displays both the sampling behavior (on the left axis) and convergence cost (on the right axis) for each method while the right column presents violin plots to visualize the convergence distribution of each method across $10$ repetitions. 

In the $3D$ \braninh~problem (see first row in Fig.\ref{fig: plots_small_noise}), all methods perform very well due to the problem's simplicity and low dimensionality. However, \cmfbo~achieves the lowest convergence cost as it leverages LF sources and employs the proposed stopping criterion. The close alignment of the mean and median of converged values with the ground truth in Fig.\ref{fig: converge_case3_01} further highlights \cmfbo's accuracy and robustness. This suggests that, on average, most repetitions converge closely to the ground truth, with minor variability introduced by a few repetitions.
Despite relying solely on HF data, \csfbo~maintains competitive convergence costs since it uses the proposed stopping condition. In comparison, qEI's robust performance is due to its excessive sampling from the HF source which, in turn, leads to much higher convergence cost. Turbo, despite sampling the same amount of HF data as qEI, struggles with noise estimation and consistently converges to values lower than the ground truth across all repetitions.

Increasing the dimension from $3D$ to $6D$ (\hartman, second row of Fig.\ref{fig: plots_small_noise}) significantly diminishes the performance of qEI which fails to converge to the ground truth in any of the $10$ repetitions. Similarly, despite having high convergence costs due to sampling the HF source $60$ times, Turbo exhibits substantial variability in its convergence values as it struggles with estimating the noise variance accurately. 
Compared to these two methods, \csfbo~ provides much lower costs on average while being quite robust to the random initializations. In this bi-fidelity example, \cmfbo~performs the best in terms of cost and is the second best (after \csfbo) considering the sensitivity to random initializations. 

We observe an interesting trend when comparing the performance of \csfbo~and \cmfbo~across \braninh~and \hartman~problems: As the input dimensionality increases, the average sampling cost increases in the case of \csfbo~while it slightly decreases for \cmfbo. We attribute this trend to the effectiveness of the LF source in effectively guiding the search process since it has a lower RRMSE in \hartman~compared to \braninh (see RRMSEs reported in Table \ref{table: analytic-formulation}).

We now investigate the results for the next two examples, \wing~and \sepwing, which differ only in the constraints of the LF sources: while the same constraint is applied to all LF sources in \wing, each LF source has its own constraint in \sepwing. This change increases the complexity in the case of \cmfbo~while it does not affect the SF cases.

The results of our studies for these two examples are illustrated in the third and fourth rows of Fig.\ref{fig: plots_small_noise}. Similar to \hartman, qEI struggles with the increased dimensionality and fails to reach the ground truth in any repetition. Its convergence values vary substantially and are quite inaccurate. In contrast to qEI, Turbo provides a more accurate and robust performance as it is a scalable method. Similar to qEI, Turbo takes $60$ HF samples before halting the optimization process and is thus expensive while it cannot exactly find the optimum in any of the $10$ repetitions. 
Comparing the three SF methods, we see that \csfbo~achieves the most robust convergence with significantly lower costs. Specifically, the short interquartile range (i.e., the length of the violin plot) as well as the close alignment of the mean and median to the ground truth, indicate that \csfbo~performs well in most repetitions.

In both \wing~and \sepwing, \cmfbo~is the most cost-effective method while maintaining accuracy. Although its reliance on LF data—rather than exclusively using HF samples like the other baselines—introduces slightly more variability in convergence, it still converges close to the ground truth in most repetitions. This is further supported by the lower median value compared to the mean, which indicates a skewed distribution. Such a pattern suggests that most of the convergence values are tightly clustered towards the optimal solution, with a few outliers that pull the mean upward.

A key difference in \cmfbo's performance between \wing~and \sepwing~is the increased complexity of learning multiple constraints in \sepwing which requires more informative sampling without incurring substantial costs. As shown by the sampling frequencies in Fig.\ref{fig: cost_wing_0.5_sep},\cmfbo~in \wing~mostly samples from LF3 which, while being moderately correlated with the HF source, is very inexpensive. However, in the case of \sepwing, \cmfbo~automatically adapts and samples more from LF1 which provides better samples than source LF3 but at a higher per-sample cost (see Table \ref{table: analytic-formulation}). We highlight that in all of our studies \cmfbo~learns the relations and relative accuracy of the LF sources and the RRMSEs in Table \ref{table: analytic-formulation} are never used by it. 

Moving on to the $20D$~example we see that the higher dimensionality combined with increased complexity adversely affects the accuracy of all methods but to different degrees. qEI consistently converges to values very far from the ground truth in all repetitions. Despite taking significantly larger number of HF samples compared to \csfbo~and \cmfbo, Turbo struggles to find the ground truth in any repetition, with errors ranging from approximately $9\%$ at best to $30\%$ at worst. \csfbo~uses fewer HF samples than Turbo and qEI (hence it has significantly lower cost) but demonstrates the least variability in convergence where its best and worst errors are reduced to approximately $4\%$ and $19\%$, respectively. 

Compared to the previous three examples, \cmfbo~leverages more HF samples than LF samples in \twenty. We attribute this behavior to the high-dimensionality of the search space and the fact that the size of the initial HF data is very small. It is also interesting to note that \cmfbo~queries LF3 and LF4 more than LF1 and LF2 which points to an attractive feature of our approach, i.e., \cmfbo~has successfully identified the most accurate and inexpensive LF sources purely based on the data (see the cost and RRMSE columns in Table \ref{table: analytic-formulation}).

Fig.\ref{fig: plots_large_noise} presents the results with large added noise which are quite similar to those in Fig.\ref{fig: plots_small_noise}. Again, the simplicity and low dimensionality of the \braninh~problem, combined with the large number of HF samples used by qEI, make it the most robust method in terms of convergence values but also the most expensive in terms of cost. 
Turbo, despite utilizing the same number of HF samples as qEI, struggles to filter out the noise and hence is sensitive to random initializations and provides inaccurate estimates in most repetitions. While the larger noise increases variability in the convergence of \cmfbo~and \csfbo, both methods perform better than qEI and Turbo. 

As the dimensionality of the problems increases, the performance of qEI deteriorates while that of Turbo increases. 
This is an expected behavior as the latter is specifically developed for high-dimensional problems. In contrast, \cmfbo~and \csfbo~not only provide smaller sampling costs, but also exhibit a robust performance as the problem dimensionality or noise increase.

Although the converged values of \csfbo~in \wing~and \sepwing (illustrated in Fig.\ref{fig: converge_wing_1} and Fig.\ref{fig: converge_wing_1_sep}) show some sensitivity to random initialization, the variations are centered around the ground truth. As expected, qEI struggles with the higher dimensionality and noise which lead to significant variability and failure in finding the ground truth in any repetition. In comparison, while using the same number of HF samples as qEI, Turbo’s scalability results in the most robust performance, though its converged values remain consistently biased and distant from the optimum in all repetitions.

In contrast to Turbo, \cmfbo~has greater variability due to its reliance on LF samples. However, it finds the exact ground truth in several repetitions and, on average, outperforms Turbo while incurring the lowest convergence cost. Additionally, a similar sampling strategy to that in Fig.\ref{fig: cost_wing_0.5_sep} is observed for \cmfbo~to manage the increased complexity of \sepwing. As shown in Fig.\ref{fig: cost_wing_1_sep}, \cmfbo~samples more from LF1 to provide more informative data to the emulator without causing substantial cost increases.

A similar trend is illustrated for \twenty, where qEI fails to find the ground truth in any repetition. As noise increases, Turbo's worst-case errors rise to approximately $40\%$, accompanied by the highest variability and also the largest convergence cost. Although the lower number of HF samples used by \csfbo~(due to the stopping criterion) and the increased noise cause slight convergence variability compared to Fig.\ref{fig: converge_20d_0.5_sep}, our method maintains its performance and its worst-case error increases by only $2\%$ compared to Fig.\ref{fig: converge_20d_0.5_sep}. Similar to Fig.~\ref{fig: cost_20d_0.5_sep}, \cmfbo~adjusts its sampling process to compensate for the higher dimensionality and increased noise in \twenty~compared to the previous benchmarks. The combination of this strategic sampling and the proposed stopping condition results in \cmfbo's lowest convergence cost and the highest robustness compared to other baselines.


    \section{Conclusions} \label{sec: conclusion}
In this paper, we introduce \cmfbo, a novel constrained MFBO framework that addresses key limitations in the existing literature. \cmfbo~effectively handles varying constraint fidelities by leveraging their underlying correlations and introduces a simple, cost-efficient constrained AF that achieves high accuracy without added complexity. Additionally, it incorporates a systematic stopping condition to automatically halt the optimization process. The effectiveness of \cmfbo~is demonstrated through analytic problems across various dimensions and noise levels.

\section{Acknowledgments}
We appreciate the support from UC National Laboratory Fees Research Program of the University of California (Grant Number $L22CR4520$). 
    \appendix
\section[Analytic Examples]{Analytic Examples}\label{appendix:a}
In this section, we provide further details on the analytic examples discussed in Section \ref{sec: results}. The functional forms of all examples and their constraints are presented in \cite{gp_plus}. We note that, in all examples, HF and LF sources are subject to different constraints. Table \ref{table: analytic-formulation} presents details on the number of initial points from each sample, the standard deviation of the added noise, the relative correlation among the HF and LF sources, and the source-dependent sampling cost. We highlight that small/large noise presented in Table \ref{table: analytic-formulation} show the standard deviation of the added noise. 

\braninh~is a bi-fidelity optimization problem in three dimensions, with $2$ constraints and the parameter space restricted to the \( [0, 1] \) domain.

\hartman, is a $6$-dimensional bi-fidelity problem with $2$ constraints and each \( x_i \in [0, 1] \). It is a popular global optimization benchmark due to its highly multi-modal nature, where the optimization algorithm must navigate complex, irregular landscapes to find the global optimum. This problem is particularly challenging for high-dimensional optimization tasks and is widely used to evaluate the scalability and robustness of optimization algorithms. 

\wing~which is designed to estimate the weight of a light aircraft which is a $10$ dimensional example, with $3$ LF sources, one constraint, and the following features: $S_{w}\in[150,200]$ is the wing area ($ft^2$), $W_{fw}\in[220, 300]$ is the weight of fuel in the wing (lb), $A \in [6, 10]$ is the aspect ratio, $\Lambda \in [-10, 10]$ is the quarter-chord sweep (deg), $\lambda \in[0.5, 1]$ is the taper ratio, $t_c \in [0.08, 0.18]$ is the aerofoil thickness to chord ratio, $N_{z} \in[2.5, 6]$ is the ultimate load factor, $W_{dg} \in[1700, 2500]$ is the flight design gross weight (lb), and $w_{p} \in[0.025, 0.08]$ is the paint weight ($lb/ft^2$). \sepwing~is similar to \wing, with one key difference: in \wing, all LF sources share the same constraint, whereas in \sepwing, each source has a distinct constraint.

\twenty~is the final problem of dimension $20$ with a single constraint, featuring $4$ LF sources. It is a mixed of polynomials of different degrees, and all its features are restricted to the \( [-0.5, 0.5] \) domain, and each fidelity source is subject to a distinct constraint (with no shared constraints between them). The formulation of these constraints is provided in \cite{gp_plus}.

    \pagebreak  
    \printbibliography
    \pagebreak
\end{document}